\RequirePackage{fix-cm}
\documentclass[twocolumn]{svjour3}    
\pdfoutput=1

\usepackage{graphicx}
\usepackage{natbib}

\usepackage{url}            
\usepackage{booktabs}       
\usepackage{amsfonts}       
\usepackage{nicefrac}       
\usepackage{microtype}      
\usepackage{graphicx}
\usepackage{enumitem}
\usepackage{bm}
\usepackage{amsmath}
\usepackage{amssymb}
\usepackage{nccmath}
\usepackage{boldline}

\usepackage{algorithm}
\usepackage[noend]{algcompatible}

\usepackage{wrapfig}
\usepackage{capt-of}
\usepackage{xcolor}
\usepackage{color,soul}

\newcommand{\nickname}{AttSets}
\newcommand{\faset}{FASet}
\newcommand{\etal}{\textit{et al}. }
\newcommand{\ie}{\textit{i}.\textit{e}., }
\newcommand{\eg}{\textit{e}.\textit{g}., }
\newcommand{\etc}{\textit{etc}. }
\newcommand{\rev}{}

\usepackage{tikz}
\newcommand*\circled[1]{\tikz[baseline=(char.base)]{
            \node[shape=circle,draw,inner sep=0.1pt] (char) {#1};}}

\begin{document}
\title{Robust Attentional Aggregation of Deep Feature Sets \\ for Multi-view 3D Reconstruction}

\author{Bo Yang  \and Sen Wang \and Andrew Markham \and Niki Trigoni }

\institute{Bo Yang, Andrew Markham, Niki Trigoni \at
         Department of Computer Science,\\ 
         University of Oxford \\ 
         \email{firstname.lastname@cs.ox.ac.uk}
         \and
         Sen Wang \at 
         School of Engineering \& Physical Sciences,\\
         Heriot-Watt University \\ 
         \email{s.wang@hw.ac.uk}
}
\date{Received: date / Accepted: date}

\maketitle

\begin{abstract}
We study the problem of recovering an underlying 3D shape from a set of images. Existing learning based approaches usually resort to recurrent neural nets, \eg GRU, or intuitive pooling operations, \eg max/mean poolings, to fuse multiple deep features encoded from input images. However, GRU based approaches are unable to consistently estimate 3D shapes given different permutations of the same set of input images as the recurrent unit is permutation variant. It is also unlikely to refine the 3D shape given more images due to the long-term memory loss of GRU. Commonly used pooling approaches are limited to capturing partial information, \eg max/mean values, ignoring other valuable features. In this paper, we present a new feed-forward neural module, named \textbf{AttSets}, together with a dedicated training algorithm, named \textbf{FASet}, to attentively aggregate an arbitrarily sized deep feature set for multi-view 3D reconstruction. The \nickname{} module is permutation invariant, computationally efficient and flexible to implement, while the \faset{} algorithm enables the \nickname{} based network to be remarkably robust and generalize to an arbitrary number of input images. We thoroughly evaluate \faset{} and the properties of \nickname{} on multiple large public datasets. Extensive experiments show that \nickname{} together with \faset{} algorithm significantly outperforms existing aggregation approaches.

\end{abstract}
\section{Introduction}
\begin{figure*}[t]
\centering
   \includegraphics[width=1\linewidth]{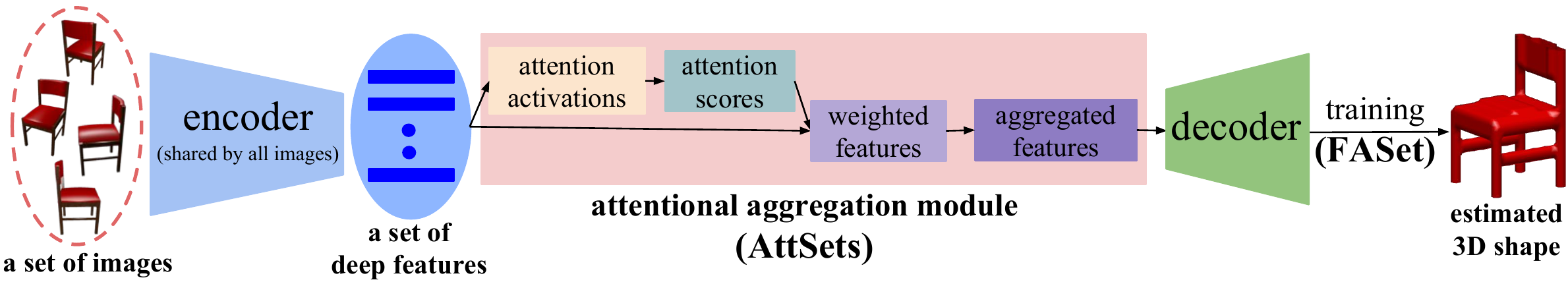}
\caption{Overview of our attentional aggregation module for multi-view 3D reconstruction. A set of $N$ images is passed through a common encoder to be a set of deep features, one element for each image. The network is trained with our \faset{} algorithm.}
\label{fig:attsets_view}
\vspace{-0.25cm}
\end{figure*}

The problem of recovering a geometric representation of the 3D world given a set of images is classically defined as multi-view 3D reconstruction in computer vision. Traditional pipelines such as Structure from Motion (SfM) \citep{Ozyesil2017} and visual Simultaneous Localization and Mapping (vSLAM) \citep{Cadena2016} typically rely on hand-crafted feature extraction and matching across multiple views to reconstruct the underlying 3D model. However, if the multiple viewpoints are separated by large baselines, it can be extremely challenging for the feature matching approach due to significant changes of appearance or self occlusions \citep{Lowe2004}. Furthermore, the reconstructed 3D shape is usually a sparse point cloud without geometric details.

Recently, a number of deep learning approaches, such as 3D-R2N2 \citep{Chan2016}, LSM \citep{Kar2017}, DeepMVS \citep{Huang2018} and RayNet \citep{Paschalidou2018} have been proposed to estimate the 3D dense shape from multiple images and have shown encouraging results. Both 3D-R2N2 \citep{Chan2016} and LSM \citep{Kar2017} formulate multi-view reconstruction as a sequence learning problem, and leverage recurrent neural networks (RNNs), particularly GRU, to fuse the multiple deep features extracted by a shared encoder from input images. However, there are three limitations. First, the recurrent network is permutation variant, \ie different permutations of the input image sequence give different reconstruction results \citep{Vinyals2016a}.
Therefore, inconsistent 3D shapes are estimated from the same image set with different permutations. Second, it is difficult to capture long-term dependencies in the sequence because of the gradient vanishing or exploding \citep{Bengio1994,Kolen2001}, so the estimated 3D shapes are unlikely to be refined even if more images are given during training and testing. Third, the RNN unit is inefficient as each element of the input sequence must be sequentially processed without parallelization \citep{Martin2018}, so is time-consuming to generate the final 3D shape given a sequence of images. 

The recent DeepMVS \citep{Huang2018} applies max pooling to aggregate deep features across a set of unordered images for multi-view stereo reconstruction, while RayNet \citep{Paschalidou2018} adopts average pooling to aggregate the deep features corresponding to the same voxel from multiple images to recover a dense 3D model. The very recent GQN \citep{Eslami2018} uses sum pooling to aggregate an arbitrary number of orderless images for 3D scene representation. Although max, average and summation poolings do not suffer from the above limitations of RNN, they tend to be `hard attentive', since they only capture the max/mean values or the summation without learning to attentively preserve the useful information. In addition, the above pooling based neural nets are usually optimized with a specific number of input images during training, therefore being not robust and general to a dynamic number of input images during testing. This critical issue is also observed in GQN \citep{Eslami2018}.

In this paper, we introduce a simple yet efficient attentional aggregation module, named \textbf{\nickname{}} \footnote{\small{Code is available at \textit{https://github.com/Yang7879/AttSets}}}. It can be easily included in an existing multi-view 3D reconstruction network to aggregate an arbitrary number of elements of a deep feature set. Inspired by the attention mechanism which shows great success in natural language processing \citep{Bahdanau2015,Raffel2016}, image captioning \citep{Xu2015b}, \textit{etc.}, we design a feed-forward neural module that can automatically learn to aggregate each element of the input deep feature set. In particular, as shown in Figure \ref{fig:attsets_view}, given a variable sized deep feature set, which are usually learnt view-invariant visual representations from a shared encoder \citep{Paschalidou2018}, our \nickname{} module firstly learns an \textbf{attention activation} for each latent feature through a standard neural layer (\eg  a fully connected layer, a 2D or 3D convolutional layer), after which an \textbf{attention score} is computed for the corresponding feature. Subsequently, the attention scores are simply multiplied by the original elements of the deep feature set, generating a set of \textbf{weighted features}. At last, the weighted features are summed across different elements of the deep feature set, producing a fixed size of \textbf{aggregated features} which are then fed into a decoder to estimate 3D shapes. Basically, this \nickname{} module can be seen as a natural extension of sum pooling into a ``weighted'' sum pooling with learnt feature-specific weights. \rev{\nickname{} shares similar concepts with the concurrent work} \citep{Ilse2018}, \rev{but it does not require the additional gating mechanism in} \citep{Ilse2018}. \rev{Notably, our simple feed-forward design allows the attention module to be separately trainable according to the property of its gradients.}

\rev{In addition, we propose a new \textbf{Feature-Attention Separate training (FASet)} algorithm that elegantly decouples the base encoder-decoder (to learn deep features) from the \nickname{} module (to learn attention scores for features). This allows the \nickname{} module to learn desired attention scores for deep feature sets and guarantees the \nickname{} based neural networks to be robust and general to dynamic sized deep feature sets.}
Basically, in the proposed training algorithm, the base encoder-decoder neural layers are only optimized when the number of input images is \textbf{1}, while the \nickname{} module is only optimized where there are more than \textbf{1} input images. Eventually, the whole optimized \nickname{} based neural network achieves superior performance with a large number of input images, while simultaneously being extremely robust and able to generalize to a small number of input images, even to a single image in the extreme case. \rev{Comparing with the widely used feed-forward attention mechanisms for visual recognition} \citep{JieHu2018,Rodriguez2018,Liu2018e,Sarafianos2018,Girdhar2017a}, \rev{our \faset{} algorithm is the first to investigate and improve the robustness of attention modules to dynamically sized input feature sets, whilst existing works are only applicable to fixed sized input data.}

\rev{Overall, our novel \nickname{} module and \faset{} algorithm are distinguished from all existing aggregation approaches in three ways. 1) Compared with RNN approaches, \nickname{} is permutation invariant and computationally efficient. 2) Compared with the widely used pooling operations, \nickname{} learns to attentively select and weight important deep features, thereby being more effective to aggregate useful information for better 3D reconstruction. 3) Compared with existing visual attention mechanisms, our \faset{} algorithm enables the whole network to be general to variable sized sets, being more robust and suitable for realistic multi-view 3D reconstruction scenarios where the number of input images usually varies dramatically.

Our key contributions are:}

\vspace{-0.15cm}
\begin{itemize}[leftmargin=0.4cm]
\item \rev{We propose an efficient feed-forward attention module, \nickname{}, to effectively aggregate deep feature sets. Our design allows the attention module to be separately optimizable according to the property of its gradients. }

\item \rev{We propose a new two-stage training algorithm, \faset{}, to decouple the base encoder/decoder and the attention module, guaranteeing the whole network to be robust and general to an arbitrary number of input images.}

\item \rev{We conduct extensive experiments on multiple public datasets, demonstrating consistent improvement over existing aggregation approaches for 3D object reconstruction from either single or multiple views.
}
\end{itemize}

\begin{figure*}[t]
\centering
   \includegraphics[width=1\linewidth]{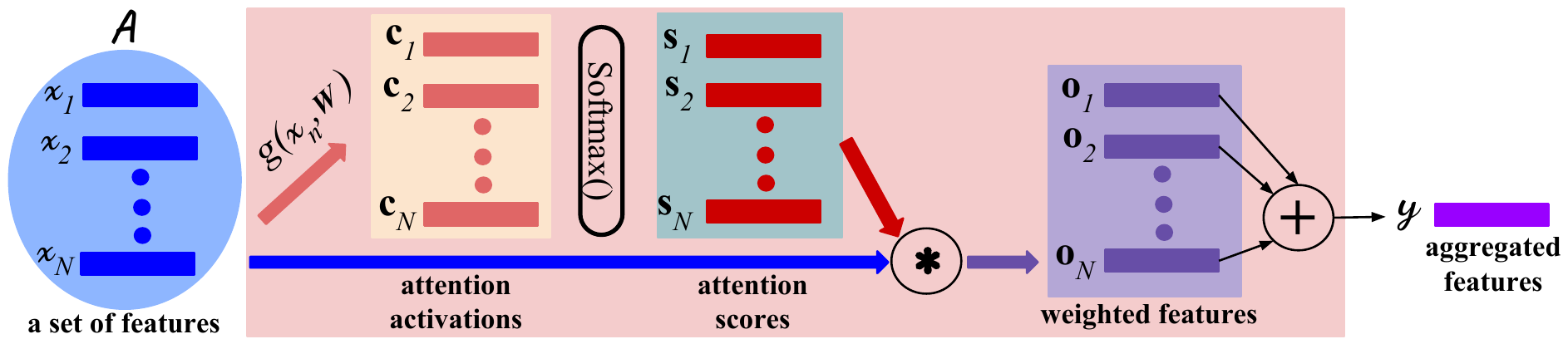}
\caption{Attentional aggregation module on sets. This module learns an attention score for each individual deep feature.}
\label{fig:attsets_f}
\vspace{-0.2cm}
\end{figure*}

\section{Related Work}
\textbf{(1) Multi-view 3D Reconstruction}. 3D shapes can be recovered from multiple color images or depth scans. To estimate the underlying 3D shape from \textbf{multiple color images}, classic SfM \citep{Ozyesil2017} and vSLAM \citep{Cadena2016} algorithms firstly extract and match hand-crafted geometric features \citep{Hartley2004} and then apply bundle adjustment \citep{Triggs2000} for both shape and camera motion estimation. Ji \etal{} \citep{Ji2017a} use ``maximizing rigidity'' for reconstruction, but this requires 2D point correspondences across images. Recent deep neural net based approaches tend to recover dense 3D shapes through learnt features from multiple images and achieve compelling results. To fuse the deep features from multiple images, both 3D-R2N2 \citep{Chan2016} and LSM \citep{Kar2017} apply the recurrent unit GRU, resulting in the networks being permutation variant and inefficient for aggregating long sequence of images. Recent SilNet \citep{Wiles2017,Wiles2018a} and DeepMVS \citep{Huang2018} simply use max pooling to preserve the first order information of multiple images, while RayNet \citep{Paschalidou2018} applies average pooling to reserve the first moment information of multiple deep features. MVSNet \citep{Yao2018} proposes a variance-based approach to capture the second moment information for multiple feature aggregation. These pooling techniques only capture partial information, ignoring the majority of the deep features. Recent SurfaceNet \citep{Ji2017b} and SuperPixel Soup \citep{Kumar2017} can reconstruct 3D shapes from two images, but they are unable to process an arbitrary number of images. As for \textbf{multiple depth image} reconstruction, the traditional volumetric fusion method \citep{Curless1996,Cao2018} integrates multiple viewpoint information by averaging truncated signed distance functions (TSDF). Recent learning based OctNetFusion \citep{Riegler2017} also adopts a similar strategy to integrate multiple depth information. However, this integration might result in information loss since TSDF values are averaged \citep{Riegler2017}. PSDF \citep{Dong2018} is recently proposed to learn a probabilistic distribution through Bayesian updating in order to fuse multiple depth images, but it is not straightforward to include the module into existing encoder-decoder networks. 

\noindent \textbf{(2) Deep Learning on Sets}. In contrast to traditional approaches operating on fixed dimensional vectors or matrices, deep learning tasks defined on sets usually require learning functions to be permutation invariant and able to process an arbitrary number of elements in a set \citep{Zaheer2017}. Such problems are widespread. Zaheer \etal introduce general permutation invariant and equivariant models in \citep{Zaheer2017}, and they end up with a \textbf{sum pooling} for permutation invariant tasks such as population statistics estimation and point cloud classification. In the very recent GQN \citep{Eslami2018}, sum pooling is also used to aggregate an arbitrary number of orderless images for 3D scene representation. Gardner \etal \citep{Gardner2017a} use \textbf{average pooling} to integrate an unordered deep feature set for classification task. Su \etal \citep{Su2015} use \textbf{max pooling} to fuse the deep feature set of multiple views for 3D shape recognition. Similarly, PointNet \citep{Qi2016} also uses max pooling to aggregate the set of features learnt from point clouds for 3D classification and segmentation. In addition, the higher-order statistics based  pooling approaches are widely used for 3D object recognition from multiple images. Vanilla \textbf{bilinear pooling} is applied for fine-grained recognition in \citep{Lin2015} and is further improved in \citep{Lin2017b}. Concurrently, \textbf{log-covariance pooling} is proposed in \citep{Ionescu2015}, and is recently generalized by \textbf{harmonized bilinear pooling} in \citep{Yu2018a}. Bilinear pooling techniques are further improved in the recent work \citep{Yu2018b,Lin2018}. However, both first-order and higher-order pooling operations ignore a majority of the information of a set. In addition, the first-order poolings do not have trainable parameters, while the higher-order poolings have only few parameters available for the network to learn. These limitations lead to the pooling based neural networks to be optimized with regards to the specific statistics of data batches during training, and therefore unable to be robust and generalize well to variable sized deep feature sets during testing. 

\noindent \textbf{(3) Attention Mechanism}. The attention mechanism was originally proposed for natural language processing \citep{Bahdanau2015}. Being coupled with RNNs, it achieves compelling results in neural machine translation \citep{Bahdanau2015}, image captioning \citep{Xu2015b}, image question answering \citep{Yang2016}, \etc However, all these coupled attention approaches are permutation variant and computationally time-consuming. Dispensing with recurrence and convolutions entirely and solely relying on attention mechanism, Transformer \citep{Vaswani2017} achieves superior performance in machine translation tasks. Similarly, being decoupled with RNNs, attention mechanisms are also applied for visual recognition \citep{JieHu2018,Rodriguez2018,Liu2018e,Sarafianos2018,Zhu2018a,Nakka2018,Girdhar2017a}, semantic segmentation \citep{Li2018a}, long sequence learning \citep{Raffel2016}, and image generation \citep{Zhang2018b}. Although the above decoupled attention modules can be used to aggregate variable sized deep feature sets, they are literally designed to operate on fixed sized features for tasks such as image recognition and generation. The robustness of attention modules regarding dynamic deep feature sets has not been investigated yet.

Compared with the original attention mechanism, our \nickname{} does not couple with RNNs. Instead, \nickname{} is a simplified feed-forward module which shares similar concepts with the concurrent work \citep{Ilse2018}. However, our \nickname{} is much simpler, without requiring the additional gating mechanism in \citep{Ilse2018}. Besides, we further propose a dedicated \faset{} algorithm, enabling the \nickname{} based network to be remarkably robust and general to arbitrarily sized deep sets. This algorithm is the first to investigate and improve the robustness of feed-forward attention mechanisms. 
\vspace{-0.25cm}

\section{\nickname{}}
\subsection{Problem Definition}
This paper considers the problem of aggregating an arbitrary number of elements of a set $\mathcal{A}$ into a fixed single output $\boldsymbol{y}$. Each element of set $\mathcal{A}$ is a feature vector extracted from a shared encoder, and the fixed dimension output $\boldsymbol{y}$ is fed into a subsequent decoder, such that the whole network can process an arbitrary number of input elements. 

Given $N$ elements in the input deep feature set $\mathcal{A} = \{\boldsymbol{x}_1, \boldsymbol{x}_2, \cdots, \boldsymbol{x}_N\}$, $\boldsymbol{x}_n \in \mathbb{R}^{1\times D}$, where $N$ is an arbitrary value, while $D$ is fixed for a specific encoder, and the output $\boldsymbol{y} \in \mathbb{R}^{1\times D}$, which is then fed into the subsequent decoder, our task is to design an aggregation function $f$ with learnable weights $\boldsymbol{\mathit{W}}$: $\boldsymbol{y} = f(\mathcal{A}, \boldsymbol{\mathit{W}})$, which should be permutation invariant, \ie for any permutation $\pi$:
\begin{equation}
f(\{ \boldsymbol{x}_1, \cdots, \boldsymbol{x}_N \}, \boldsymbol{\mathit{W}}) = f(\{\boldsymbol{x}_{\pi(1)}, \cdots, \boldsymbol{x}_{\pi(N)} \}, \boldsymbol{\mathit{W}})
\end{equation}
The common pooling operations, \eg max/mean/sum, are the simplest instantiations of function $f$ where $\boldsymbol{\mathit{W}} \in \emptyset$. However, these pooling operations are predefined to capture partial information.

\subsection{\nickname{} Module} \label{sec:design}
\begin{figure*}[t]
\setlength{\abovecaptionskip}{ 1 pt}
\setlength{\belowcaptionskip}{ 1 pt}
\centering
   \includegraphics[width=1.0\linewidth]{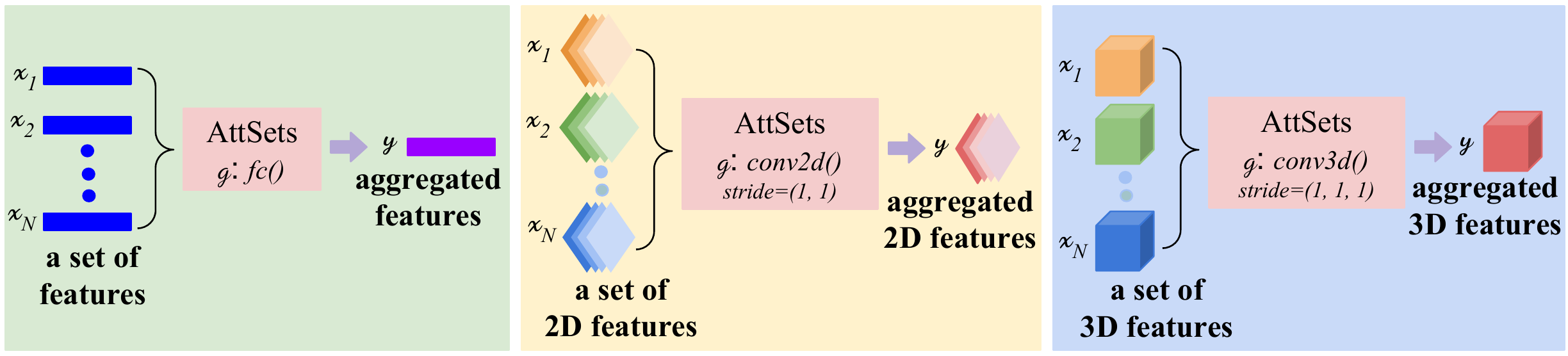}
\caption{Implementation of \nickname{} with fully connected layer, 2D ConvNet, and 3D ConvNet. These three variants of \nickname{} can be flexibly plugged into different locations of an existing encoder-decoder network.}
\label{fig:attsets_imp}
\vspace{-0.1cm}
\end{figure*}

The basic idea of our \nickname{} module is to learn an attention score for each latent feature of the whole deep feature set. In this paper, each latent feature refers to each entry of an individual element of the feature set, with an individual element usually represented by a latent vector, \ie $\boldsymbol{x}_n$. The learnt scores can be regarded as a mask that automatically selects useful latent features across the set. The selected features are then summed across multiple elements of the set.

As shown in Figure \ref{fig:attsets_f}, given a set of features $\mathcal{A} = \{\boldsymbol{x}_1, \boldsymbol{x}_2, \cdots, \boldsymbol{x}_N\}$, $\boldsymbol{x}_n \in \mathbb{R}^{1\times D}$, \nickname{} aims to fuse it into a fixed dimensional output $\boldsymbol{y}$, where $\boldsymbol{y} \in \mathbb{R}^{1\times D}$.

To build the \nickname{} module, we first feed each element of the feature set $\mathcal{A}$ into a shared function $g$ which can be a standard neural layer, \ie a linear transformation layer without any non-linear activation functions. Here we use a fully connected layer as an example, the bias term is dropped for simplicity. The output of function $g$ is a set of learnt attention activations $\mathcal{C}=\{\boldsymbol{c}_1, \boldsymbol{c}_2, \cdots, \boldsymbol{c}_N\}$, where
\begin{equation}
\begin{gathered}
\boldsymbol{c}_n = g(\boldsymbol{x}_n, \boldsymbol{W}) = \boldsymbol{x}_n\boldsymbol{W},\\ 
(\boldsymbol{x}_n \in \mathbb{R}^{1\times D}, \quad \boldsymbol{W} \in \mathbb{R}^{D\times D}, \quad \boldsymbol{c}_n \in \mathbb{R}^{1\times D} )
\end{gathered}
\end{equation}

Secondly, the learnt attention activations are normalized across the $N$ elements of the set, computing a set of attention scores $\mathcal{S}=\{\boldsymbol{s}_1, \boldsymbol{s}_2, \cdots, \boldsymbol{s}_N\}$. We choose $softmax$ as the normalization operation, so the attention scores for the $n^{th}$ feature element are
\begin{equation}
\begin{gathered}
\boldsymbol{s}_n = [s^1_n, s^2_n, \cdots, s^d_n, \cdots, s^D_n],\\
s^d_n = \frac{e^{c^d_n}}{\sum^N_{j=1}{ e^{c^d_j}}}, \textit{$c^d_n$, $c^d_j$ are the $d^{th}$ entry of $\boldsymbol{c}_n$, $\boldsymbol{c}_j$.}
\end{gathered}
\end{equation}

Thirdly, the computed attention scores $\mathcal{S}$ are multiplied by their corresponding original feature set $\mathcal{A}$, generating a new set of deep features, denoted as weighted features $\mathcal{O}=\{\boldsymbol{o}_1, \boldsymbol{o}_2, \cdots, \boldsymbol{o}_N\}$, where
\begin{ceqn}
\begin{align}
\boldsymbol{o}_n = \boldsymbol{x}_n * \boldsymbol{s}_n
\end{align}
\end{ceqn}

Lastly, the set of weighted features $\mathcal{O}$ are summed up across the total $N$ elements to get a fixed size feature vector, denoted as $\boldsymbol{y}$, where
\begin{equation}
\begin{gathered}
\boldsymbol{y} =[y^1, y^2, \cdots, y^d, \cdots, y^D],\\
y^d = \sum^N_{n=1}o^d_n, \qquad \textit{$o^d_n$ is the $d^{th}$ entry of $\boldsymbol{o}_n$.}
\end{gathered}
\end{equation}

In the above formulation, we show how \nickname{} gradually aggregates a set of $N$ feature vectors $\mathcal{A}$ into a single vector $\boldsymbol{y}$, where $\boldsymbol{y} \in \mathbb{R}^{1\times D}$. 

\subsection{Permutation Invariance}
The output of \nickname{} module $\boldsymbol{y}$ is permutation invariant with regard to the input deep feature set $\mathcal{A}$. Here is the simple proof. 
\begin{equation}
\setlength{\abovedisplayskip}{3pt}
\setlength{\belowdisplayskip}{0pt}
\label{eq:yd}
[y^1, \cdots y^d \cdots, y^D] = f(\{\boldsymbol{x}_1, \cdots \boldsymbol{x}_n \cdots, \boldsymbol{x}_N\}, \boldsymbol{W})
\end{equation}

In Equation \ref{eq:yd}, the $d^{th}$ entry of the output $\boldsymbol{y}$ is computed as follows:
\begin{ceqn}
\begin{align}
\label{eq:yd2}
y^d &= \sum^N_{n=1}o^d_n  
= \sum^N_{n=1}(x^d_n*s^d_n) \nonumber \\
&= \sum^N_{n=1}\left( x^d_n * \frac{e^{c^d_n}}{\sum^N_{j=1} e^{c^d_j}  }  \right) \nonumber \\
&= \sum^N_{n=1}\left( x^d_n * \frac{e^{(\boldsymbol{x}_n\boldsymbol{w}^d)} } 
{ \sum^N_{j=1}e^{(\boldsymbol{x}_j\boldsymbol{w}^d})} \right)  \nonumber \\
&=\frac{\sum^N_{n=1} \left( x^d_n * e^ {(\boldsymbol{x}_n\boldsymbol{w}^d)} \right) }
{\sum^N_{j=1}e^{(\boldsymbol{x}_j\boldsymbol{w}^d)}}, \qquad 
\end{align}
\end{ceqn}
\textit{where} $\boldsymbol{w}^d$ is the $d^{th}$ column of the weights $\boldsymbol{W}$. In above Equation \ref{eq:yd2}, both the denominator and numerator are a summation of a permutation equivariant term. Therefore the value $y^d$, and also the full vector $\boldsymbol{y}$, is invariant to different permutations of the deep feature set $\mathcal{A}=\{\boldsymbol{x}_1, \boldsymbol{x}_2, \cdots, \boldsymbol{x}_n, \cdots, \boldsymbol{x}_N\}$ \citep{Zaheer2017}. 

\subsection{Implementation}\label{sec:impl}
In Section \ref{sec:design}, we described how our \nickname{} aggregates an arbitrary number of vector features into a single vector, where the attention activation learning function $g$ embeds a fully connected ($fc$) layer. \nickname{} can also be easily implemented with both 2D and 3D convolutional neural layers to aggregate both 2D and 3D deep feature sets, thus being flexible to be included into a 2D encoder/decoder or 3D encoder/decoder. Particularly, as shown in Figure \ref{fig:attsets_imp}, to aggregate a set of 2D features, \ie a tensor of $(width\times height\times channels)$, the attention activation learning function $g$ embeds a standard $conv2d$ layer with a stride of $(1\times 1)$. Similarly, to fuse a set of 3D features, \ie a tensor of $(width\times height\times depth\times channels)$, the function $g$ embeds a standard $conv3d$ layer with a stride of $(1\times 1\times 1)$. For the above $conv2d$/$conv3d$ layer, the filter size can be 1, 3 or many. The larger the filter size, the learnt attention score is considered to be correlated with the larger local spatial area.

Instead of embedding a single neural layer, the function $g$ is also flexible to include multiple layers, but the tensor shape of the output of function $g$ is required to be consistent with the input element $\boldsymbol{x}_n$. This guarantees each individual feature of the input set $\mathcal{A}$ will be associated with a learnt and unique weight. For example, a standard 2-layer or 3-layer ResNet module \citep{He2016b} could be a candidate of the function $g$. The more layers that $g$ embeds, the capability of \nickname{} module is expected to increase accordingly.

Compared with $fc$ enabled \nickname{}, the $conv2d$ or $conv3d$ based \nickname{} variants tend to have fewer learnable parameters. Note that both the $conv2d$ and $conv3d$ based \nickname{} are still permutation invariant, as the function $g$ is shared across all elements of the deep feature set and it does not depend on the order of the elements \citep{Zaheer2017}.

\section{FASet}
\subsection{Motivation}\label{sec:optim_motiv}
\begin{algorithm*}[h]
\caption{\small Feature-Attention Separate training of an \nickname{} enabled network. $M$ is batch size, $N$ is image number.
}
\label{alg:jtso}
\begin{algorithmic} 
\vspace{2mm}
\STATE{\textbf{Stage 1:}}
\FOR{number of training iterations}{}
    \STATE{$\bullet$ Sample $M$ sets of images $\{ \mathcal{I}_1, \cdots, \mathcal{I}_m, \cdots, \mathcal{I}_M \}$ and sample $N$ images for each set, \ie $\mathcal{I}_m = \{\boldsymbol{i}_m^1, \cdots, \boldsymbol{i}_m^n, \cdots, \boldsymbol{i}_m^N \}$. 
 Sample $M$ 3D shape labels $\{ \boldsymbol{v}_1, \cdots, \boldsymbol{v}_m, \cdots, \boldsymbol{v}_M \}$.}
    
    \vspace{2mm}
    \STATE{$\bullet$ Update the base network by ascending its stochastic gradient:\\ 
    \vspace{-5.2mm}
        \[
        	\nabla_{\Theta_{base}} \frac{1}{MN} \sum_{m=1}^M \sum_{n=1}^N \left[ \ell(\boldsymbol{\hat{v}}_m^n, \boldsymbol{v}_m) \right], \text{ $where$ $\boldsymbol{\hat{v}}_m^n$ is the estimated 3D shape of single image $\{\boldsymbol{i}_m^n$\}.}
        \]
        }
    \ENDFOR

\STATE{\textbf{Stage 2:}}
\FOR{number of training iterations}{}
    \STATE{$\bullet$ Sample $M$ sets of images $\{ \mathcal{I}_1, \cdots, \mathcal{I}_m, \cdots, \mathcal{I}_M \}$ and sample $N$ images for each set, \ie $\mathcal{I}_m = \{\boldsymbol{i}_m^1, \cdots, \boldsymbol{i}_m^n, \cdots, \boldsymbol{i}_m^N \}$. 
 Sample $M$ 3D shape labels $\{ \boldsymbol{v}_1, \cdots, \boldsymbol{v}_m, \cdots, \boldsymbol{v}_M \}$.}
 
   \vspace{2mm}
    \STATE{$\bullet$ Update the \nickname{} module by ascending its stochastic gradient:\\
        \vspace{-5mm}
        \[
        	\nabla_{\Theta_{att}} \frac{1}{M} \sum_{m=1}^M \left[ \ell(\boldsymbol{\hat{v}}_m, \boldsymbol{v}_m) \right],         \text{$where$ $\boldsymbol{\hat{v}}_m$ is the estimated 3D shape of the image set $\mathcal{I}_m$.}
        \]
        }
\ENDFOR
  
  \vspace{-1mm}
  \\The gradient-based updates can use any gradient optimization algorithm.
\end{algorithmic}
\end{algorithm*}

Our \nickname{} module can be included in an existing encoder-decoder multi-view 3D reconstruction network, replacing the RNN units or pooling operations. Basically, in an \nickname{} enabled encoder-decoder net, the encoder-decoder serves as the base architecture to learn visual features for shape estimation, while the \nickname{} module learns to assign different attention scores to combine those features. As such, the base network tends to have robustness and generality with regard to different input image content, while the \nickname{} module tends to be general regarding an arbitrary number of input images. 

However, to achieve this robustness is not straightforward. The standard end-to-end joint optimization approach is unable to guarantee that the base encoder-decoder and \nickname{} are able to learn visual features and the corresponding scores separately, because there are no explicit feature score labels available to directly supervise the \nickname{} module. 

Let us revisit the previous Equation \ref{eq:yd2} as follows and draw insights from it.
\begin{ceqn}
\begin{align}
\label{eq:yd2_2}
y^d =\frac{\sum^N_{n=1} \left( x^d_n * e^ {(\boldsymbol{x}_n\boldsymbol{w}^d)} \right) }
{\sum^N_{j=1}e^{(\boldsymbol{x}_j\boldsymbol{w}^d)}}
\end{align}
\end{ceqn}
\textit{where} $N$ is the size of an arbitrary input set and $\boldsymbol{w}^d$ are the \nickname{} parameters to be optimized. If $N$ is 1, then the equation can be simplified as
\begin{ceqn}
\begin{align}
\label{eq:yd2_3}
y^d &= x^d_n \\
\frac{\partial y^d}{\partial x^d_n} =1, \qquad   \label{eq:yd2_33}
\frac{\partial y^d}{\partial \boldsymbol{w}^d} &=\boldsymbol{0}, \qquad N=1 
\end{align}
\end{ceqn}
This shows that all parameters, \ie $\boldsymbol{w}^d$, of the \nickname{} module are not going to be optimized when the size of the input feature set is 1. 

However, if $N>1$, Equation \ref{eq:yd2_2} is unable to be simplified to Equation \ref{eq:yd2_3}. Therefore, 
\begin{ceqn}
\begin{align}
\label{eq:yd2_44}
\frac{\partial y^d}{\partial x^d_n} \neq 1, \qquad
\frac{\partial y^d}{\partial \boldsymbol{w}^d } \neq \boldsymbol{0}, \qquad N>1
\end{align}
\end{ceqn}
This shows that both the parameters of \nickname{} and the base encoder-decoder layers will be optimized simultaneously, if the whole network is trained in the standard end-to-end fashion.

Here arises the critical issue. When $N>1$, all derivatives of the parameters in the \textbf{encoder} are different from the derivatives when $N=1$ due to the chain rule of differentiation applied backwards from $\frac{\partial y^d}{\partial x^d_n}$. \rev{Put simply, the derivatives of encoder are \textit{$N$-dependent}. As a consequence, the encoded visual features and the learnt attention scores would be \textit{$N$-biased} if the whole network is jointly trained. This biased network is unable to generalize to an arbitrary value of $N$ during testing.}

\rev{To illustrate the above issue, assuming the base encoder-decoder and the \nickname{} module are jointly trained given $5$ images to reconstruct every object, the base encoder will be eventually optimized towards $5$-view object reconstruction during training.} The trained network can indeed perform well given 5 views during testing, but it is unable to predict a satisfactory object shape given only 1 image.

To alleviate the above problem, a naive approach is to enumerate various values of $N$ during the jointly training, such that the final optimized network can be somehow robust and general to arbitrary $N$ during testing. However, this approach would inevitably optimize the encoder to learn the \textit{mean} features of input data for varying $N$. The overall performance will hence not be optimal. In addition, it is impractical and also time-consuming to enumerate all values of $N$ during training.

\vspace{-0.45cm}
\subsection{Algorithm}
To resolve the critical issue discussed in Section \ref{sec:optim_motiv}, we propose a \textbf{Feature-Attention Separate training (FASet)} algorithm that decouples the base encoder-decoder and the \nickname{} module, enabling the base encoder-decoder to learn robust deep features and the \nickname{} module to learn the desired attention scores for the feature sets.

In particular, the base encoder-decoder neural layers are only optimized when the number of input images is 1, while the \nickname{} module is only optimized where there are more than 1 input images. In this regard, the parameters of the base encoding layers would have consistent derivatives in the whole training stage, thus being optimized steadily. In the meantime, the \nickname{} module would be optimized solely based on multiple elements of learnt visual features from the shared encoder. 

The trainable parameters of the base encoder-decoder are denoted as $\boldsymbol{\Theta}_{base}$, and the trainable parameters of \nickname{} module are denoted as $\boldsymbol{\Theta}_{att}$, and the loss function of the whole network is represented by $\ell$ which is determined by the specific supervision signal of the base network. Our \faset{} is shown in Algorithm \ref{alg:jtso}. It can be seen that $\boldsymbol{\Theta}_{base}$ and $\boldsymbol{\Theta}_{att}$ are completely decoupled from one another, thus being separately optimized in two stages. In stage 1, the $\boldsymbol{\Theta}_{base}$ is firstly well optimized until convergence, which guarantees the base encoder-decoder is able to learn robust and general visual features. In stage 2, the $\boldsymbol{\Theta}_{att}$ is optimized to learn attention scores for individual visual features. Basically, this module learns to select and weight important deep features automatically. 

\rev{In \faset{} algorithm, once the} $\boldsymbol{\Theta}_{base}$ \rev{is well optimized in stage 1, it is not necessary to train it again, since the two-stage algorithm guarantees that optimizing} $\boldsymbol{\Theta}_{base}$ \rev{is agnostic to the attention module. The \faset{} algorithm is a crucial component to maintain the superior robustness of the \nickname{} module, as shown in Section} \ref{sec:sig_faset}. \rev{Without it, the feed-forward attention mechanism is ineffective with respect to dynamic input sets.}

\section{Evaluation}
\begin{figure}[t]
\centering
   \includegraphics[width=1\linewidth]{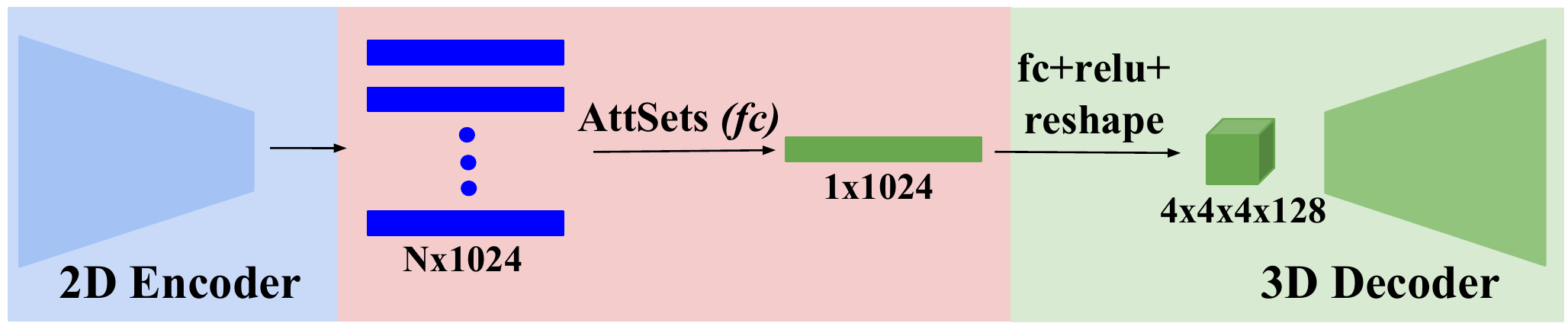}
\caption{The architecture of Base$_{\textrm{r2n2}}$-AttSets for multi-view 3D reconstruction network. The base encoder-decoder is the same as 3D-R2N2.}
\label{fig:fig_attsets_base_r2n2_fc}
\vspace{-0.15cm}
\end{figure}

\begin{figure}[t]
\centering
   \includegraphics[width=1\linewidth]{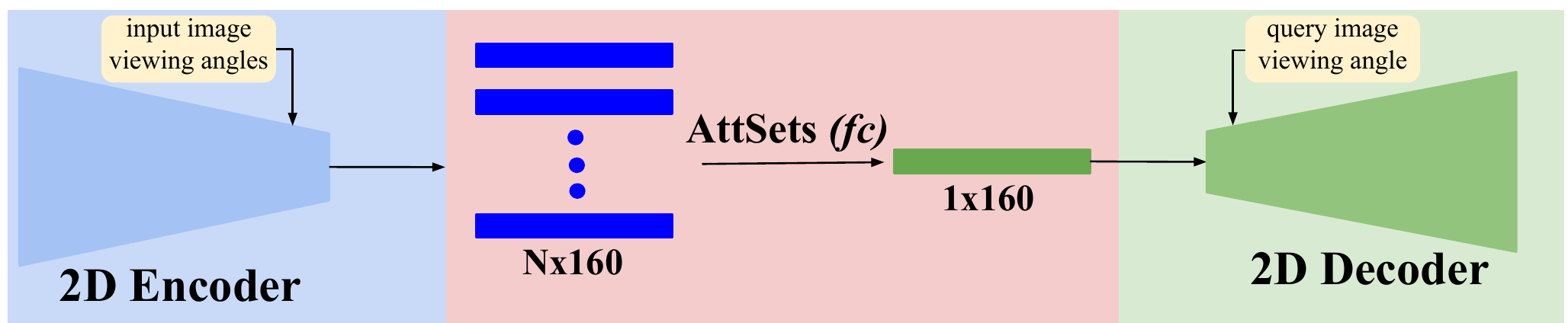}
\caption{The architecture of Base$_{\textrm{silnet}}$-AttSets for multi-view 3D shape learning. The base encoder-decoder is the same as SilNet.}
\label{fig:fig_attsets_base_silnet_fc}
\vspace{-0.35cm}
\end{figure}

\textbf{Base Networks.} To evaluate the performance and various properties of \nickname{}, we choose the encoder-decoders of 3D-R2N2 \citep{Chan2016} and SilNet \citep{Wiles2017} as two base networks. 
\begin{itemize}[leftmargin=0.25cm]
\item Encoder-decoder of 3D-R2N2. The original 3D-R2N2 consists of (1) a shared ResNet-based 2D encoder which encodes a size of $127\times 127 \times 3$ images into $1024$ dimensional latent vectors, (2) a GRU module which fuses $N$ $1024$ dimensional latent vectors into a single $4\times 4\times 4\times 128$ tensor, and (3) a ResNet-based 3D decoder which decodes the single tensor into a $32\times 32\times 32$ voxel grid representing the 3D shape. Figure \ref{fig:fig_attsets_base_r2n2_fc} shows the architecture of \nickname{} based multi-view 3D reconstruction network where the only difference is that the original GRU module is replaced by \nickname{} in the middle. This network is called Base$_{\textrm{r2n2}}$-AttSets.

\item Encoder-decoder of SilNet. The original SilNet consists of (1) a shared 2D encoder which encodes a size of $127\times 127\times 3$ images together with image viewing angles into $160$ dimensional latent vectors, (2) a max pooling module which aggregates $N$ latent vectors into a single vector, and (3) a 2D decoder which estimates an object silhouette ($57\times 57$) from the single latent vector and a new viewing angle. Instead of being explicitly supervised by 3D shape labels, SilNet aims to implicitly learn a 3D shape representation from multiple images via the supervision of 2D silhouettes. Figure \ref{fig:fig_attsets_base_silnet_fc} shows the architecture of \nickname{} based SilNet where the only difference is that the original max pooling is replaced by \nickname{} in the middle. This network is called Base$_{\textrm{silnet}}$-AttSets.

\end{itemize}
\begin{table*}[t]
\caption{ Group 1: mean IoU for multi-view reconstruction of all 13 categories in ShapeNet$_{\textrm{r2n2}}$ testing split. All networks are firstly trained given only 1 image for each object in Stage 1. The \nickname{} module is further trained given \textbf{2 images} per object in Stage 2, while other competing approaches are fine-tuned given \textbf{2 images} per object in Stage 2.}
\centering
\label{tab:iou_r2n2_02v}
\tabcolsep=0.125cm
\begin{tabular}{ l|cccccccccc}
\hline
&1 view&2 views&3 views& 4 views&5 views&8 views&12 views&16 views&20 views&24 views \\
\hline
Base$_{\textrm{r2n2}}$-GRU &0.574&0.608&0.622&0.629&0.633&0.639&0.642&0.642&0.641&0.640 \\
Base$_{\textrm{r2n2}}$-max pooling &0.620&0.651&0.660&0.665&0.666&0.671&0.672&0.674&0.673&0.673 \\
Base$_{\textrm{r2n2}}$-mean pooling &0.632&0.654&0.661&0.666&0.667&0.674&0.676&0.680&0.680&0.681 \\
Base$_{\textrm{r2n2}}$-sum pooling &0.633&0.657&0.665&0.669&0.669&0.670&0.666&0.667&0.666&0.665 \\
Base$_{\textrm{r2n2}}$-BP pooling &0.588&0.608&0.615&0.620&0.621&0.627&0.628&0.632&0.633&0.633 \\
Base$_{\textrm{r2n2}}$-MHBN pooling &0.578&0.599&0.606&0.611&0.612&0.618&0.620&0.623&0.624&0.624 \\
Base$_{\textrm{r2n2}}$-SMSO pooling &0.623&0.654&0.664&0.670&0.672&0.679&0.679&0.682&0.680&0.678 \\
\textbf{Base$_{\textrm{r2n2}}$-\nickname{}(Ours)} &\textbf{0.642}&\textbf{0.665}&\textbf{0.672}&\textbf{0.677}&\textbf{0.678}&\textbf{0.684}
&\textbf{0.686}&\textbf{0.690}&\textbf{0.690}&\textbf{0.690} \\
\hline
\end{tabular}
\vspace{-0.1 cm}
\end{table*}

\begin{table*}[t]
\caption{Group 2: mean IoU for multi-view reconstruction of all 13 categories in ShapeNet$_{\textrm{r2n2}}$ testing split. All networks are firstly trained given only 1 image for each object in Stage 1. The \nickname{} module is further trained given \textbf{8 images} per object in Stage 2, while other competing approaches are fine-tuned given \textbf{8 images} per object in Stage 2.}
\centering
\label{tab:iou_r2n2_08v}
\tabcolsep=0.125cm
\begin{tabular}{ l|cccccccccc}
\hline
&1 view&2 views&3 views& 4 views&5 views&8 views&12 views&16 views&20 views&24 views \\
\hline
Base$_{\textrm{r2n2}}$-GRU &0.580&0.616&0.629&0.637&0.641&0.649&0.652&0.652&0.652&0.652 \\
Base$_{\textrm{r2n2}}$-max pooling &0.524&0.615&0.641&0.655&0.661&0.674&0.678&0.683&0.684&0.684 \\
Base$_{\textrm{r2n2}}$-mean pooling &0.632&0.657&0.665&0.670&0.672&0.679&0.681&0.685&0.686&0.686 \\
Base$_{\textrm{r2n2}}$-sum pooling &0.580&0.628&0.644&0.656&0.660&0.672&0.677&0.682&0.684&0.685 \\
Base$_{\textrm{r2n2}}$-BP pooling &0.544&0.599&0.618&0.628&0.632&0.644&0.648&0.654&0.655&0.656 \\
Base$_{\textrm{r2n2}}$-MHBN pooling &0.570&0.596&0.606&0.612&0.614&0.621&0.624&0.628&0.629&0.629 \\
Base$_{\textrm{r2n2}}$-SMSO pooling &0.570&0.621&0.641&0.652&0.656&0.668&0.673&0.679&0.680&0.681 \\
\textbf{Base$_{\textrm{r2n2}}$-\nickname{}(Ours)} &\textbf{0.642}&\textbf{0.662}&\textbf{0.671}&\textbf{0.676}&\textbf{0.678}&\textbf{0.686}
&\textbf{0.688}&\textbf{0.693}&\textbf{0.694}&\textbf{0.694} \\
\hline
\end{tabular}
\vspace{-0.1 cm}
\end{table*}

\begin{table*}[t]
\caption{Group 3: mean IoU for multi-view reconstruction of all 13 categories in ShapeNet$_{\textrm{r2n2}}$ testing split. All networks are firstly trained given only 1 image for each object in Stage 1. The \nickname{} module is further trained given \textbf{16 images} per object in Stage 2, while other competing approaches are fine-tuned given \textbf{16 images} per object in Stage 2.}
\centering
\label{tab:iou_r2n2_16v}
\tabcolsep=0.125cm
\begin{tabular}{ l|cccccccccc}
\hline
&1 view&2 views&3 views& 4 views&5 views&8 views&12 views&16 views&20 views&24 views \\
\hline
Base$_{\textrm{r2n2}}$-GRU &0.579&0.614&0.628&0.636&0.640&0.647&0.651&0.652&0.653&0.653 \\
Base$_{\textrm{r2n2}}$-max pooling &0.511&0.604&0.633&0.649&0.656&0.671&0.678&0.684&0.686&0.686 \\
Base$_{\textrm{r2n2}}$-mean pooling &0.594&0.637&0.652&0.662&0.667&0.677&0.682&0.687&0.688&0.689 \\
Base$_{\textrm{r2n2}}$-sum pooling &0.570&0.629&0.647&0.657&0.664&0.678&0.684&0.690&0.692&0.692 \\
Base$_{\textrm{r2n2}}$-BP pooling &0.545&0.593&0.611&0.621&0.627&0.637&0.642&0.647&0.648&0.649 \\
Base$_{\textrm{r2n2}}$-MHBN pooling &0.570&0.596&0.606&0.612&0.614&0.621&0.624&0.627&0.628&0.629 \\
Base$_{\textrm{r2n2}}$-SMSO pooling &0.580&0.627&0.643&0.652&0.656&0.668&0.673&0.679&0.680&0.681 \\
\textbf{Base$_{\textrm{r2n2}}$-\nickname{}(Ours)} &\textbf{0.642}&\textbf{0.660}&\textbf{0.668}&\textbf{0.673}&\textbf{0.676}&\textbf{0.683}
&\textbf{0.687}&\textbf{0.691}&\textbf{0.692}&\textbf{0.693} \\
\hline
\end{tabular}
\vspace{-0.1 cm}
\end{table*}
\begin{table*}[t]
\caption{Group 4: mean IoU for multi-view reconstruction of all 13 categories in ShapeNet$_{\textrm{r2n2}}$ testing split. All networks are firstly trained given only 1 image for each object in Stage 1. The \nickname{} module is further trained given \textbf{24 images} per object in Stage 2, while other competing approaches are fine-tuned given \textbf{24 images} per object in Stage 2.}
\centering
\label{tab:iou_r2n2_24v}
\tabcolsep=0.125cm
\begin{tabular}{ l|cccccccccc}
\hline
&1 view&2 views&3 views& 4 views&5 views&8 views&12 views&16 views&20 views&24 views \\
\hline
Base$_{\textrm{r2n2}}$-GRU &0.578 &0.613&0.627&0.635&0.639&0.647&0.651&0.653&0.653&0.654 \\
Base$_{\textrm{r2n2}}$-max pooling &0.504&0.600&0.631&0.648&0.655&0.671&0.679&0.685&0.688&0.689 \\
Base$_{\textrm{r2n2}}$-mean pooling &0.593&0.634&0.649&0.659&0.663&0.673&0.677&0.683&0.684&0.685 \\
Base$_{\textrm{r2n2}}$-sum pooling &0.580&0.634&0.650&0.658&0.660&0.678&0.682&0.689&0.690&0.691 \\
Base$_{\textrm{r2n2}}$-BP pooling &0.524&0.585&0.609&0.623&0.630&0.644&0.650&0.656&0.659&0.660 \\
Base$_{\textrm{r2n2}}$-MHBN pooling &0.566&0.595&0.606&0.613&0.616&0.624&0.627&0.631&0.632&0.632 \\
Base$_{\textrm{r2n2}}$-SMSO pooling &0.556&0.613&0.635&0.647&0.653&0.668&0.674&0.681&0.682&0.684 \\
\textbf{Base$_{\textrm{r2n2}}$-\nickname{}(Ours)} &\textbf{0.642}&\textbf{0.660}&\textbf{0.668}&\textbf{0.674}&\textbf{0.676}&\textbf{0.684}
&\textbf{0.688}&\textbf{0.693}&\textbf{0.694}&\textbf{0.695} \\
\hline
\end{tabular}
\vspace{-0.1 cm}
\end{table*}

\begin{table*}[t]
\caption{Group 5: mean IoU for multi-view reconstruction of all 13 categories in ShapeNet$_{\textrm{r2n2}}$ testing split. All networks are firstly trained given only 1 image for each object in Stage 1. The \nickname{} module is further trained given random number of images per object in Stage 2, \ie $N$ is uniformly sampled from \textbf{[1, 24]}, while other competing approaches are fine-tuned given random number of views per object in Stage 2.}
\centering
\label{tab:iou_r2n2_allv}
\tabcolsep=0.125cm
\begin{tabular}{ l|cccccccccc}
\hline
&1 view&2 views&3 views& 4 views&5 views&8 views&12 views&16 views&20 views&24 views \\
\hline
Base$_{\textrm{r2n2}}$-GRU &0.580&0.615&0.629&0.637&0.641&0.648&0.651&0.651&0.651&0.651 \\
Base$_{\textrm{r2n2}}$-max pooling &0.601&0.638&0.652&0.660&0.663&0.673&0.677&0.682&0.683&0.684 \\
Base$_{\textrm{r2n2}}$-mean pooling &0.598&0.637&0.652&0.660&0.664&0.675&0.679&0.684&0.685&0.686 \\
Base$_{\textrm{r2n2}}$-sum pooling &0.587&0.631&0.646&0.656&0.660&0.672&0.678&0.683&0.684&0.685 \\
Base$_{\textrm{r2n2}}$-BP pooling &0.582&0.610&0.620&0.626&0.628&0.635&0.638&0.641&0.642&0.643 \\
Base$_{\textrm{r2n2}}$-MHBN pooling &0.575&0.599&0.608&0.613&0.615&0.622&0.624&0.628&0.629&0.629 \\
Base$_{\textrm{r2n2}}$-SMSO pooling &0.580&0.624&0.641&0.652&0.657&0.669&0.674&0.679&0.681&0.682 \\
\textbf{Base$_{\textrm{r2n2}}$-\nickname{}(Ours)} &\textbf{0.642}&\textbf{0.662}&\textbf{0.670}&\textbf{0.675}&\textbf{0.677}&\textbf{0.685}
&\textbf{0.688}&\textbf{0.692}&\textbf{0.693}&\textbf{0.694} \\
\hline
\end{tabular}
\vspace{-0.1 cm}
\end{table*}

\textbf{Competing Approaches.} We compare our \nickname{} and \faset{} with three groups of competing approaches. Note that all the following competing approaches are connected at the same location of the base encoder-decoder shown in the pink block of Figure \ref{fig:fig_attsets_base_r2n2_fc} and \ref{fig:fig_attsets_base_silnet_fc}, with the same network configurations and training/testing settings.

\begin{itemize}[leftmargin=0.3cm]
\item RNNs. The original 3D-R2N2 makes use of the \textbf{GRU} \citep{Chan2016,Kar2017} unit for feature aggregation and serves as a solid baseline.
\item First-order poolings. The widely used \textbf{max}/\textbf{mean}/ \textbf{sum} pooling operations \citep{Huang2018,Paschalidou2018,Eslami2018} are all implemented for comparison.
\item Higher-order poolings. We also compare with the state-of-the-art higher-order pooling approaches, including bilinear pooling (\textbf{BP}) \citep{Lin2015}, and the very recent \textbf{MHBN} \citep{Yu2018a} and \textbf{SMSO} poolings \citep{Yu2018b}. 
\end{itemize}

\textbf{Datasets.} All approaches are evaluated on four large open datasets.
\begin{itemize} [leftmargin=0.3cm]
\item ShapeNet$_{\textrm{r2n2}}$ Dataset \citep{Chan2016}. The released 3D-R2N2 dataset consists of $13$ categories of $43,783$ common objects with synthesized RGB images from the large scale ShapeNet 3D repository \citep{Chang2015}. For each 3D object, $24$ images are rendered from different viewing angles circling around each object. The train/test dataset split is $0.8:0.2$.
\item ShapeNet$_{\textrm{lsm}}$ Dataset \citep{Kar2017}. Both LSM and 3D-R2N2 datasets are generated from the same 3D ShapeNet repository \citep{Chang2015}, \ie they have the same ground truth labels regarding the same object. However, the ShapeNet$_{\textrm{lsm}}$ dataset has totally different camera viewing angles and lighting sources for the rendered RGB images. Therefore, we use the ShapeNet$_{\textrm{lsm}}$ dataset to evaluate the robustness and generality of all approaches. All images of ShapeNet$_{\textrm{lsm}}$ dataset are resized from $224\times 224$ to $127\times 127$ through linear interpolation.
\item ModelNet40 Dataset. ModelNet40 \citep{Wu2015} consists of $12,311$ objects belonging to 40 categories. The 3D models are split into $9,843$ training samples and $2,468$ testing samples. For each 3D model, it is voxelized into a $30\times30\times30$ shape in \citep{Qi2016a}, and 12 RGB images are rendered from different viewing angles. 
All 3D shapes are zero-padded to be $32\times32\times32$, and the images are linearly resized from $224\times224$ to $127\times127$ for training and testing.
\item Blobby Dataset \citep{Wiles2017}. It contains $11,706$ blobby objects. Each object has $5$ RGB images paired with viewing angles and the corresponding silhouettes, which are generated from Cycles in Blender under different lighting sources and texture models. 
\end{itemize}

\textbf{Metrics.} The explicit 3D voxel reconstruction performance of Base$_{\textrm{r2n2}}$-AttSets and the competing approaches is evaluated on three datasets: ShapeNet$_{\textrm{r2n2}}$, ShapeNet$_{\textrm{lsm}}$ and ModelNet40. We use the mean Intersection-over-Union (IoU) \citep{Chan2016} between predicted 3D voxel grids and their ground truth as the metric. The IoU for an individual voxel grid is formally defined as follows:
\begin{ceqn}
\begin{align*}
IoU = \frac{\sum_{i=1}^{L} \left[  I (h_i>p) * I(\bar{h_i}) \right] }{ \sum_{i=1}^{L}   \left[I  \left( I(h_{i} >p) + I(\bar{h_i}) \right) \right] } 
\end{align*}
\end{ceqn}
$where$ $I(\cdot)$ is an indicator function, $h_{i}$ is the predicted value for the $i^{th}$ voxel, $\bar{h_i}$ is the corresponding ground truth, $p$ is the threshold for voxelization, $L$ is the total number of voxels in a whole voxel grid. As there is no validation split in the above three datasets, to calculate the IoU scores, we independently search the optimal binarization threshold value from $0.2 \sim 0.8$ with a step $0.05$ for all approaches for fair comparison. In our experiments, we found that all optimal thresholds of different approaches end up with $0.3$ or $0.35$.

The implicit 3D shape learning performance of Base$_{\textrm{silnet}}$-AttSets and the competing approaches is evaluated on the Blobby dataset. The mean IoU between predicted 2D silhouettes and the ground truth is used as the metric \citep{Wiles2017}.

\subsection{Evaluation on ShapeNet$_{\textrm{r2n2}}$ Dataset}\label{sec:eval_r2n2}
To fully evaluate the aggregation performance and robustness, we train the Base$_{\textrm{r2n2}}$-AttSets and its competing approaches on ShapeNet$_{\textrm{r2n2}}$ dataset. For fair comparison, all networks \rev{(the pooling/GRU/\nickname{} based approaches)} are trained according to the proposed two-stage training algorithm.

\textbf{Training Stage 1.} All networks are trained given only 1 image for each object, \ie $N=1$ in all training iterations, until convergence. Basically, this is to guarantee all networks are well optimized for the extreme case where there is only one input image.

\textbf{Training Stage 2.} To enable these networks to be more robust for multiple input images, all networks are further trained given more images per object. Particularly, we conduct the following five parallel groups of training experiments.

\begin{figure*}[!h]
	\setlength{\abovecaptionskip}{0 cm}
	\setlength{\belowcaptionskip}{-8pt}
	\begin{minipage}[t]{0.19\textwidth}
		\centerline{
			\includegraphics[width=1\textwidth]{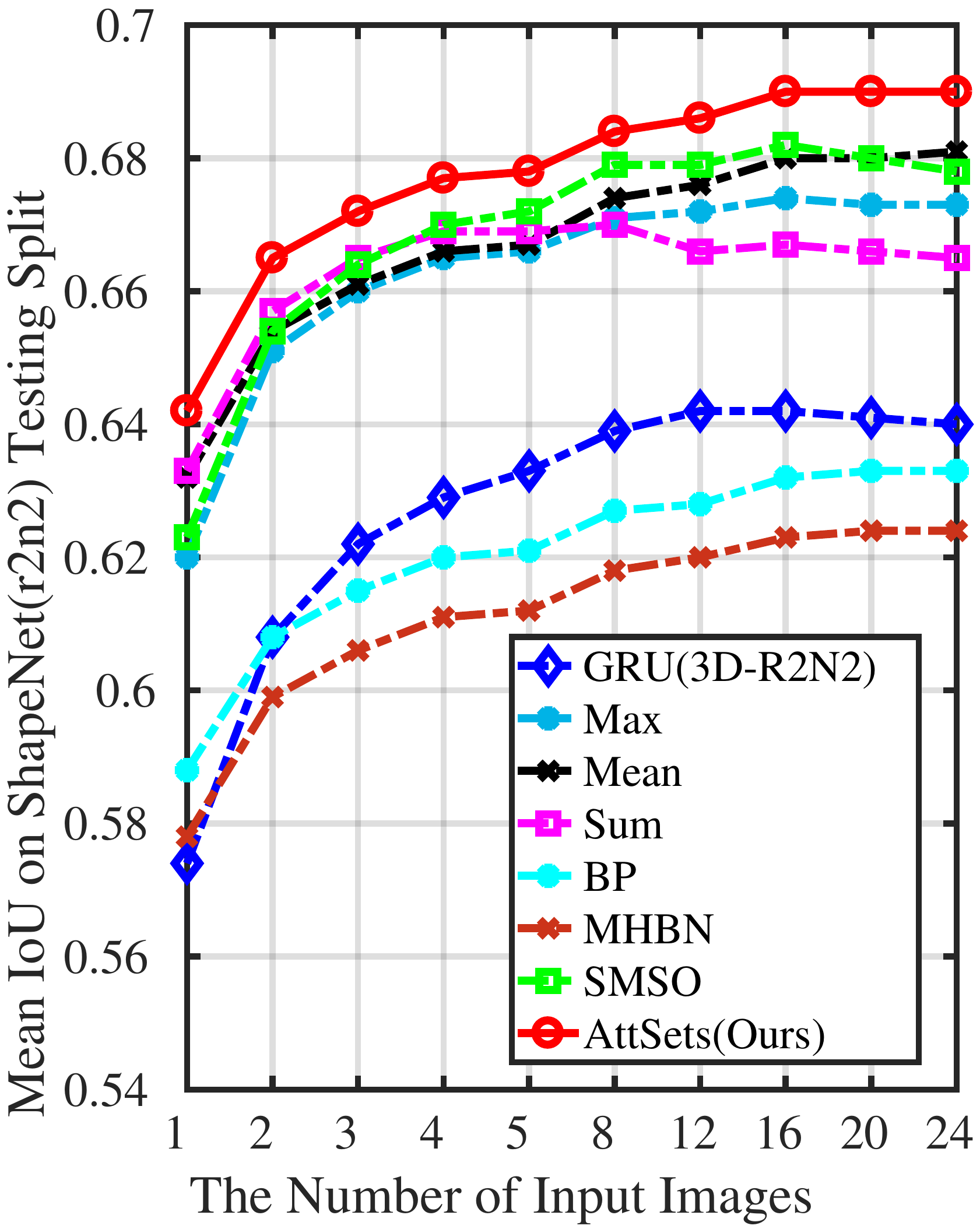}}
		\caption{IoUs of Group 1.}
		\label{fig:02v_mIoU}
	\end{minipage}
	\makebox[0.02in][]{}
	\begin{minipage}[t]{0.19\textwidth}
		\centerline{
			\includegraphics[width=1.0\textwidth]{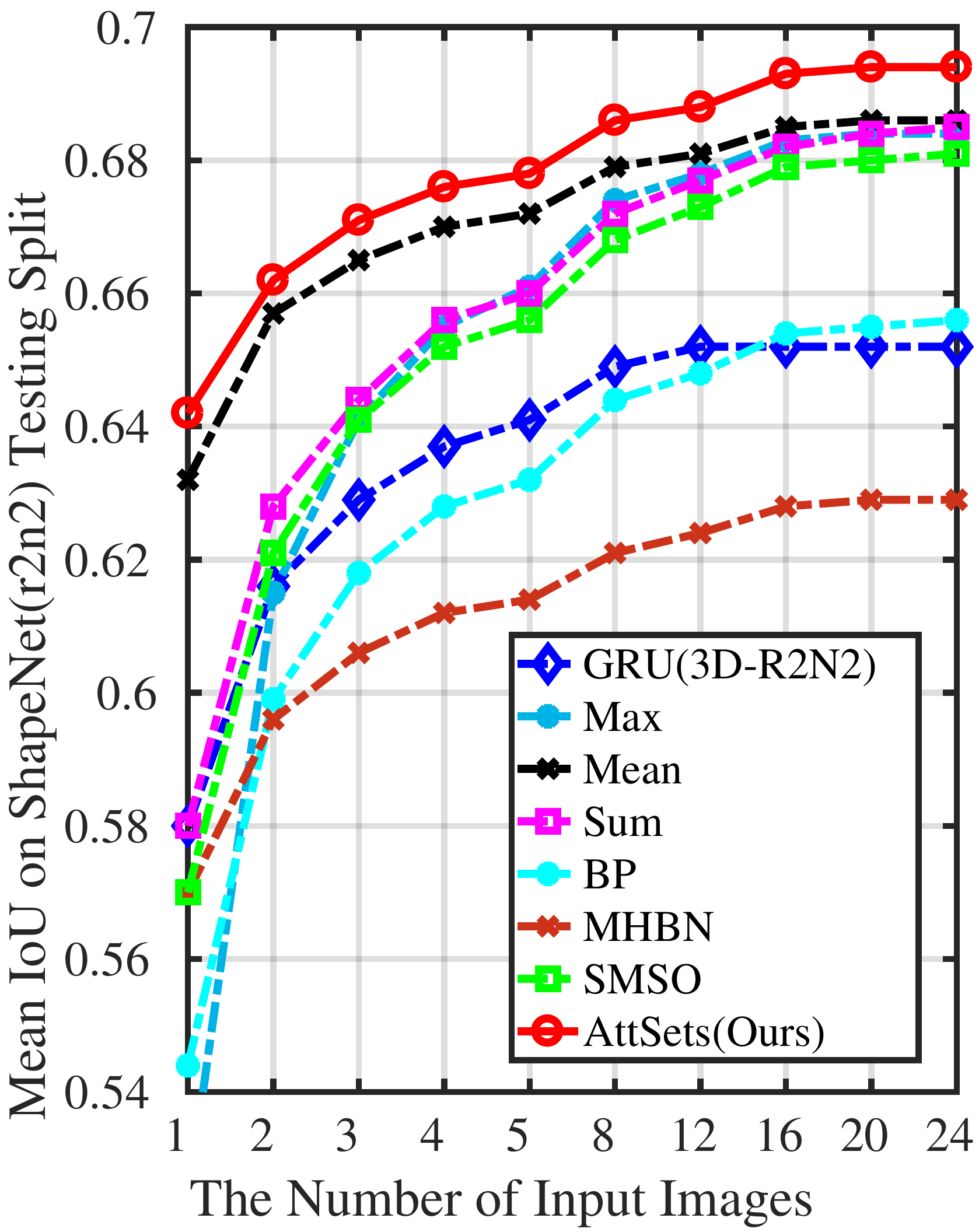}}
		\caption{IoUs of Group 2.}
        \label{fig:08v_mIoU}
	\end{minipage}
	\makebox[0.02in][]{}
	\begin{minipage}[t]{0.19\textwidth}
		\centerline{
			\includegraphics[width=1.0\textwidth]{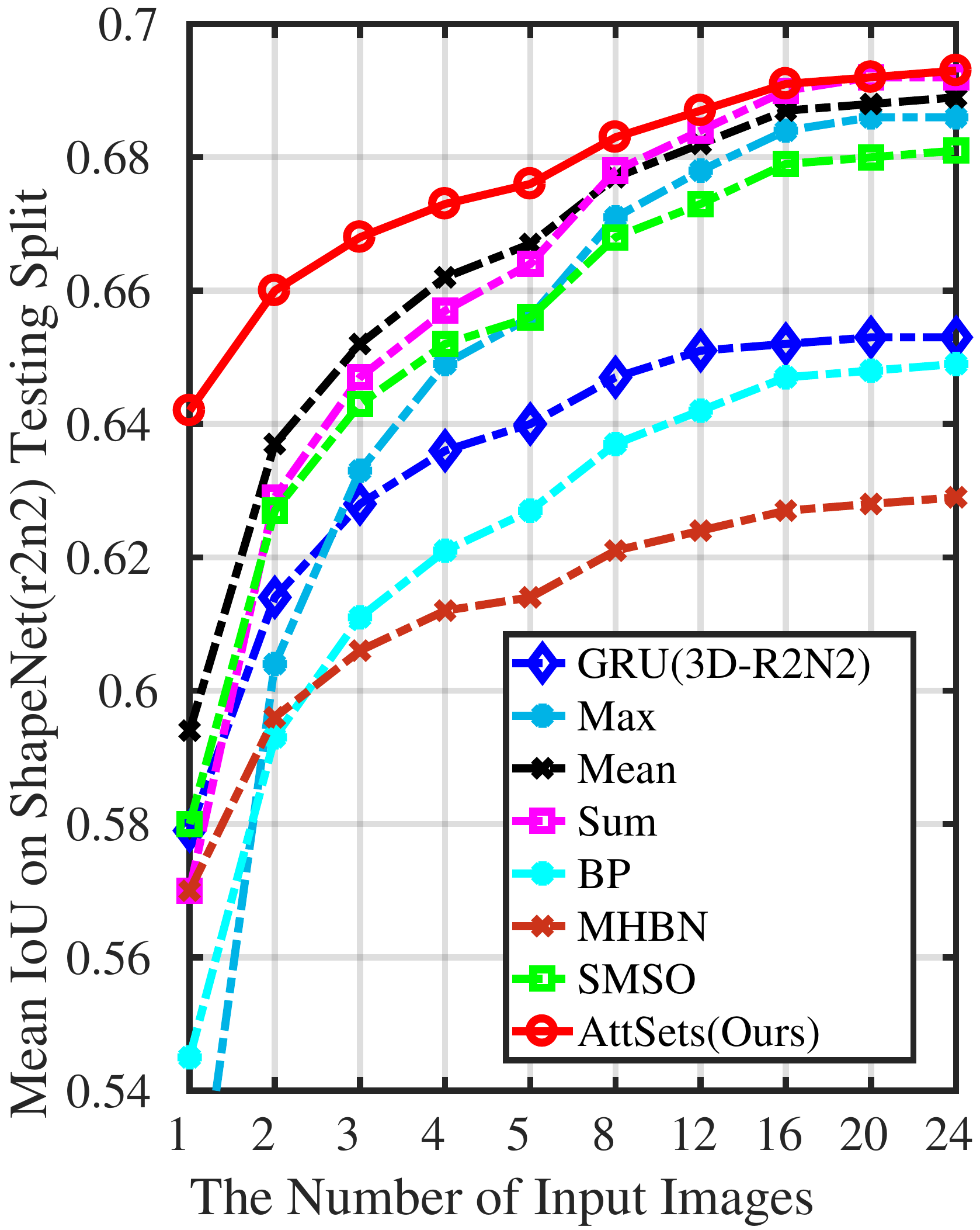}}
		\caption{IoUs of Group 3.}
        \label{fig:16v_mIoU}
	\end{minipage}
    \makebox[0.02in][]{}
	\begin{minipage}[t]{0.19\textwidth}
		\centerline{
			\includegraphics[width=1\textwidth]{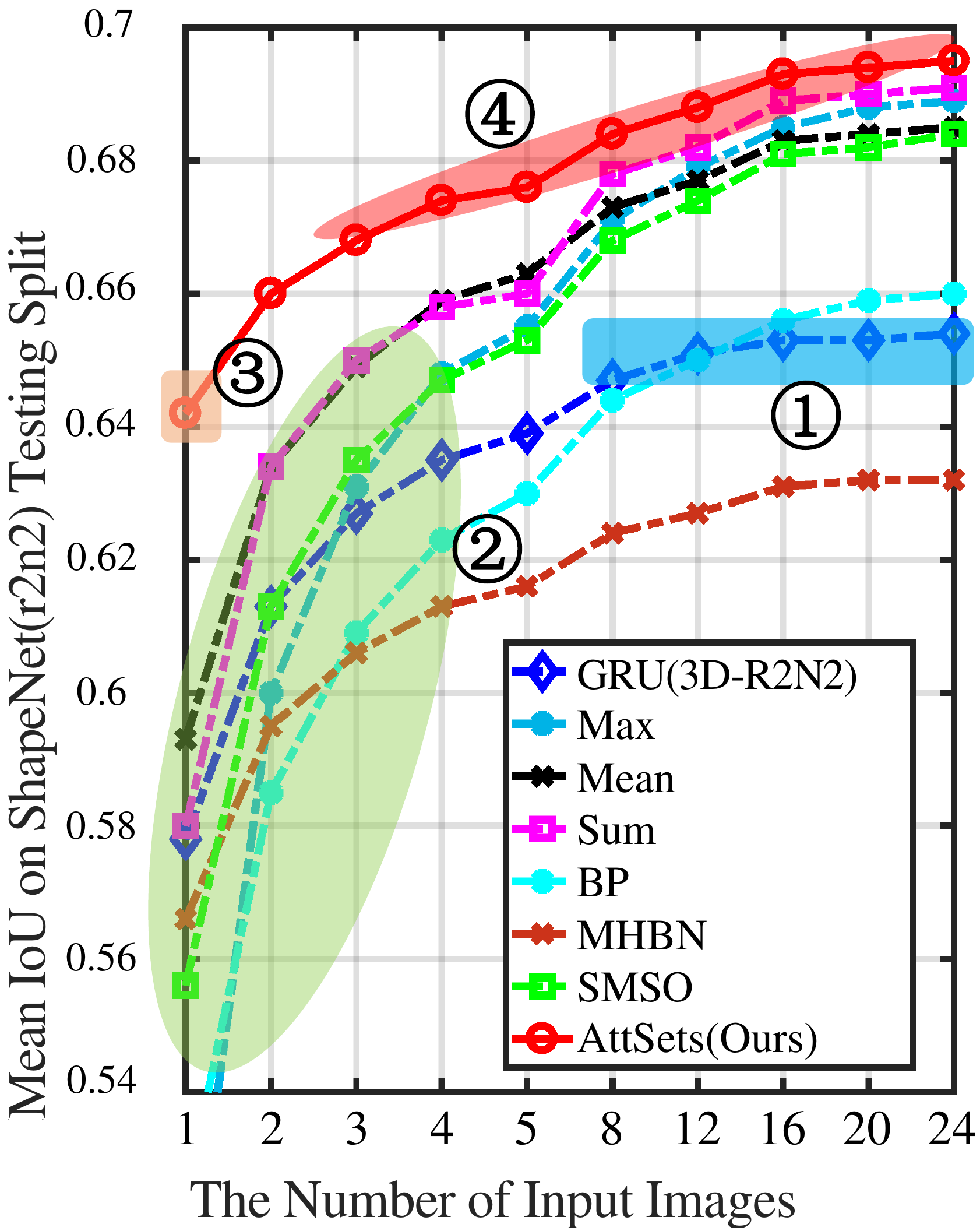}}
		\caption{IoUs of Group 4.}
        \label{fig:24v_mIoU}
	\end{minipage}
	\makebox[0.02in][]{}
	\begin{minipage}[t]{0.19\textwidth}
		\centerline{
			\includegraphics[width=1.0\textwidth]{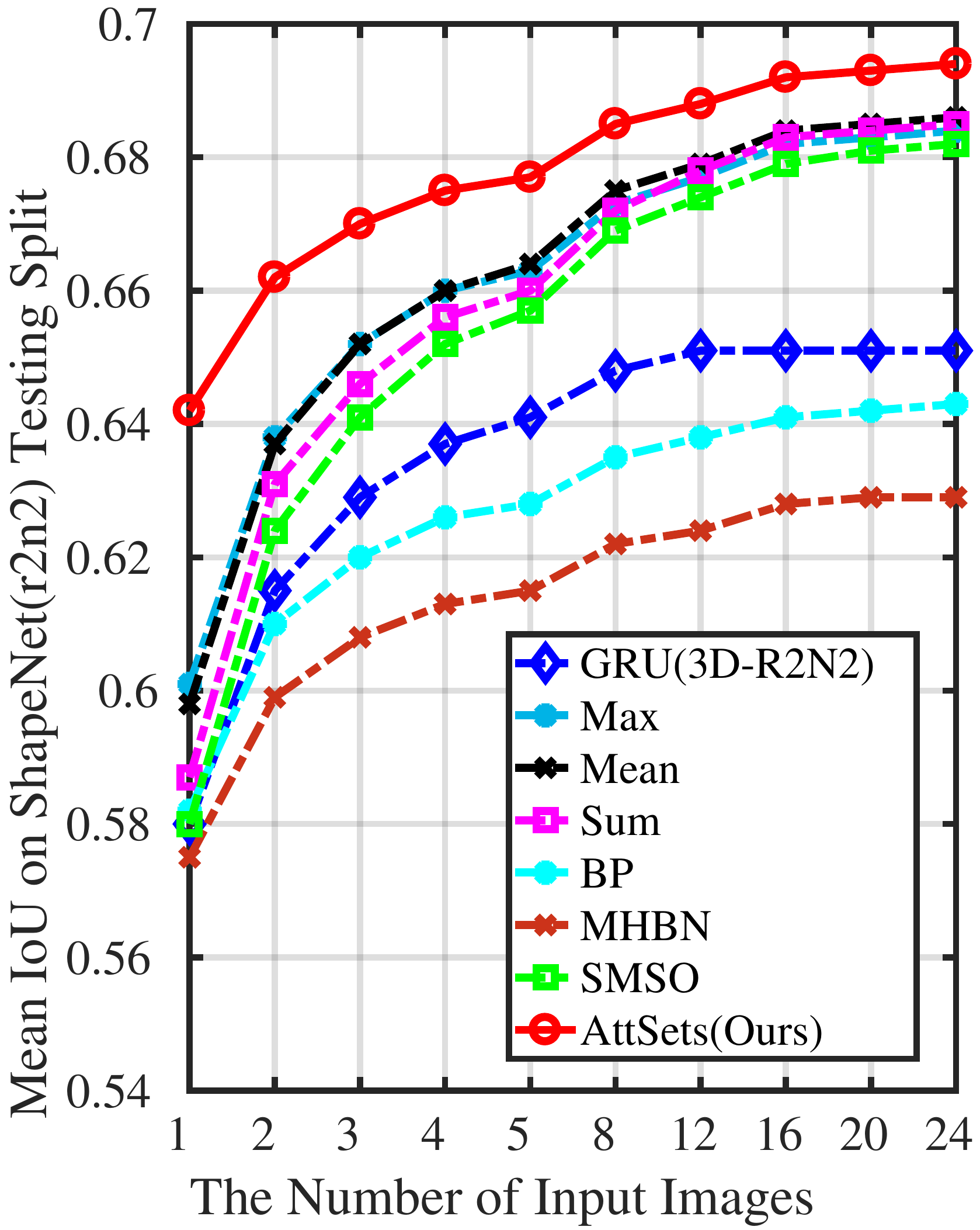}}
		\caption{\scriptsize{IoUs of Group} 5.}
        \label{fig:allv_mIoU}
	\end{minipage}
\vspace{-0.4cm}
\end{figure*}

\begin{figure*}[t]
\centering
   \includegraphics[width=1\linewidth]{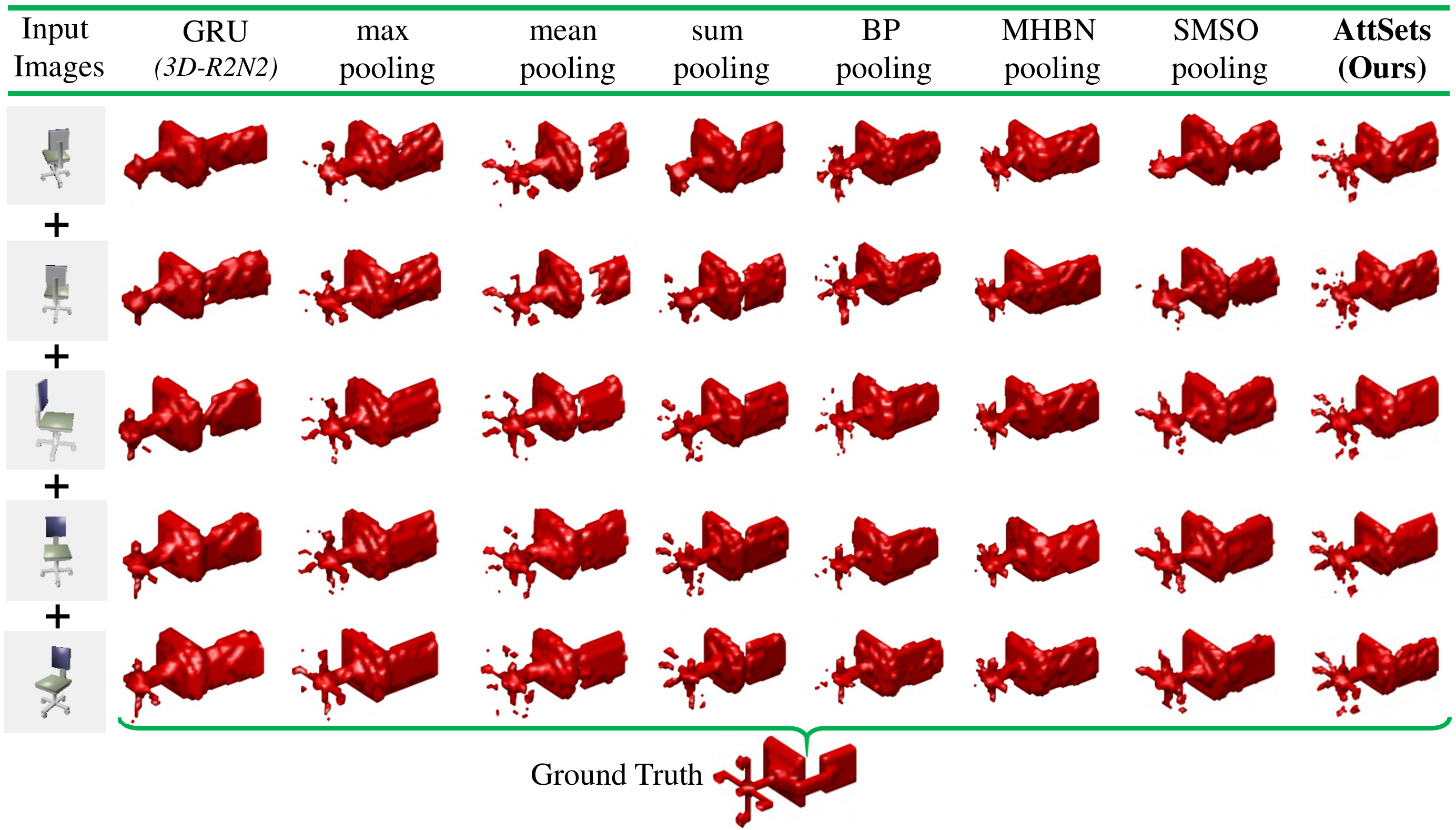}
\caption{Qualitative results of multi-view reconstruction from different approaches in experiment Group 5.}
\label{fig:mv_demo_r2n2}
\vspace{-0.35cm}
\end{figure*}
\begin{table*}[t]
\caption{ Per-category mean IoU for single view reconstruction on ShapeNet$_{\textrm{r2n2}}$ testing split.}
\centering
\label{tab:iou_r2n2_1v}
\tabcolsep=0.02cm
\begin{tabular}{ l|cccccccccccccc}
\hline
&plane&bench&cabinet&car&chair&monitor&lamp&speaker&firearm&couch&table&phone&watercraft&mean \\
\hline
Base$_{\textrm{r2n2}}$-GRU &0.530&0.449&0.730&0.807&0.487&0.497&0.391&0.671&0.553&0.631&0.515&0.683&0.535&0.580\\
Base$_{\textrm{r2n2}}$-max pooling &0.573&0.521&0.755&0.835&0.533&0.544&0.423&0.695&0.587&0.678&0.562&0.710&0.582&0.620\\
Base$_{\textrm{r2n2}}$-mean pooling &0.582&0.540&0.773&0.837&0.547&0.550&0.440&0.713&0.595&0.695&0.576&0.718&0.593&0.632\\
Base$_{\textrm{r2n2}}$-sum pooling &0.588&0.536&0.771&0.838&0.554&0.547&0.442&0.710&0.598&0.690&0.575&0.728&0.598&0.633\\
Base$_{\textrm{r2n2}}$-BP pooling &0.536&0.469&0.747&0.816&0.484&0.499&0.398&0.678&0.556&0.646&0.528&0.681&0.550&0.588\\
Base$_{\textrm{r2n2}}$-MHBN pooling &0.528&0.451&0.742&0.812&0.471&0.487&0.386&0.677&0.548&0.637&0.515&0.674&0.546&0.578\\
Base$_{\textrm{r2n2}}$-SMSO pooling &0.572&0.521&0.763&0.833&0.541&0.548&0.433&0.704&0.581&0.682&0.566&0.721&0.581&0.623\\
OGN &0.587&0.481&0.729&0.816&0.483&0.502&0.398&0.637&0.593&0.646&0.536&0.702&\textbf{0.632}&0.596\\
AORM &\textbf{0.605}&0.498&0.715&0.757&0.532&0.524&0.415&0.623&\textbf{0.618}&0.679&0.547&0.738&0.552&0.600\\
PointSet &0.601&0.550&0.771&0.831&0.544&0.552&\textbf{0.462}&\textbf{0.737}&0.604&\textbf{0.708}&\textbf{0.606}&\textbf{0.749}&0.611&0.640\\
\textbf{Base$_{\textrm{r2n2}}$-\nickname{}\scriptsize{(Ours)}} &0.594&\textbf{0.552}&\textbf{0.783}&\textbf{0.844}&\textbf{0.559}&\textbf{0.565}
&0.445&0.721&0.601&0.703&0.590&0.743&0.601&\textbf{0.642}\\
\hline
\end{tabular}
\vspace{-.25 cm}
\end{table*}

\begin{itemize}[leftmargin=0.3cm]
\item Group 1. All networks are further trained given only 2 images for each object, \ie $N=2$ in all iterations. As to our Base$_{\textrm{r2n2}}$-AttSets, the well-trained encoder-decoder in previous stage 1 is frozen, and we only optimize the \nickname{} module according to our \faset{} algorithm \ref{alg:jtso}. As to the competing approaches, \eg GRU and all poolings, we turn to fine-tune the whole networks because they do not have separate parameters suitable for special training. To be specific, we use smaller learning rate (1e-5) to carefully train these networks to achieve better performance where $N=2$ until convergence.

\item Group 2/3/4. Similarly, in these three groups of second-stage training experiments, $N$ is set to be 8, 16, 24 separately. 

\item Group 5. All networks are further trained until convergence, but $N$ is uniformly and randomly sampled from $[1, 24]$ for each object during training. In the above Group 1/2/3/4, $N$ is fixed for each object, while $N$ is dynamic for each object in this Group 5. 
\end{itemize}

The above experiment Groups 1/2/3/4 are designed to investigate how all competing approaches would be further optimized towards the statistics of a fixed $N$ during training, thus resulting in different level of robustness given an arbitrary number of $N$ during testing. By contrast, the paradigm in Group 5 aims at enumerating all possible $N$ values during training. Therefore the overall performance might be more robust regarding an arbitrary number of input images during testing, compared with the above Group 1/2/3/4 experiments. 

\textbf{Testing Stage.}
All networks trained in above five groups of experiments are separately tested given $N = \{1, 2, 3, 4, 5, 8, 12, 16, 20,24\}$. The permutations of input images are the same for all different approaches for fair comparison. Note that, we do not test the networks which are only trained in Stage 1, because the \nickname{} module is not optimized and the corresponding Base$_{\textrm{r2n2}}$-AttSets is unable to generalize to multiple input images during testing. Therefore, it is meaningless to compare the performance when the network is solely trained on a single image.

\textbf{Results.} Tables \ref{tab:iou_r2n2_02v} $\sim$ \ref{tab:iou_r2n2_allv} show the mean IoU scores of all 13 categories for experiments of Group 1 $\sim$ 5, while Figures \ref{fig:02v_mIoU} $\sim$ \ref{fig:allv_mIoU} show the trends of mean IoU changes in different Groups. Figure \ref{fig:mv_demo_r2n2} shows the estimated 3D shapes in experiment Group 5, with an increasing number of images from 1 to 5 for different approaches. 

We notice that the reported IoU scores of ShapeNet data repository in original LSM \citep{Kar2017} are higher than our scores. However, the experimental settings in LSM \citep{Kar2017} are quite different from ours in the following two aspects. 1) The original LSM requires both RGB images and the corresponding viewing angles as input, while all our experiments do not. 2) The original LSM dataset has different styles of rendered color images and different train/test splits compared with our experimental settings. Therefore the reported IoU scores in LSM are not directly comparable with ours and we do not include the results in this paper to avoid confusion. Note that, the aggregation module of LSM \citep{Kar2017}, \ie GRU, is the same as used in 3D-R2N2 \citep{Chan2016}, and is indeed fully evaluated throughout our experiments.

To highlight the performance of single view 3D reconstruction, Table \ref{tab:iou_r2n2_1v} shows the optimal per-category IoU scores for different competing approaches from experiments Group 1 $\sim$ 5. In addition, we also compare with the state-of-the-art dedicated single view reconstruction approaches including OGN \citep{Tatarchenko2017}, AORM \citep{Yang2018c} and PointSet \citep{Fan2017} in Table \ref{tab:iou_r2n2_1v}. Overall, our \nickname{} based approach outperforms all others by a large margin for either single view or multi view reconstruction, and generates much more compelling 3D shapes.

\begin{figure*}[h]
\centering
   \includegraphics[width=1\linewidth]{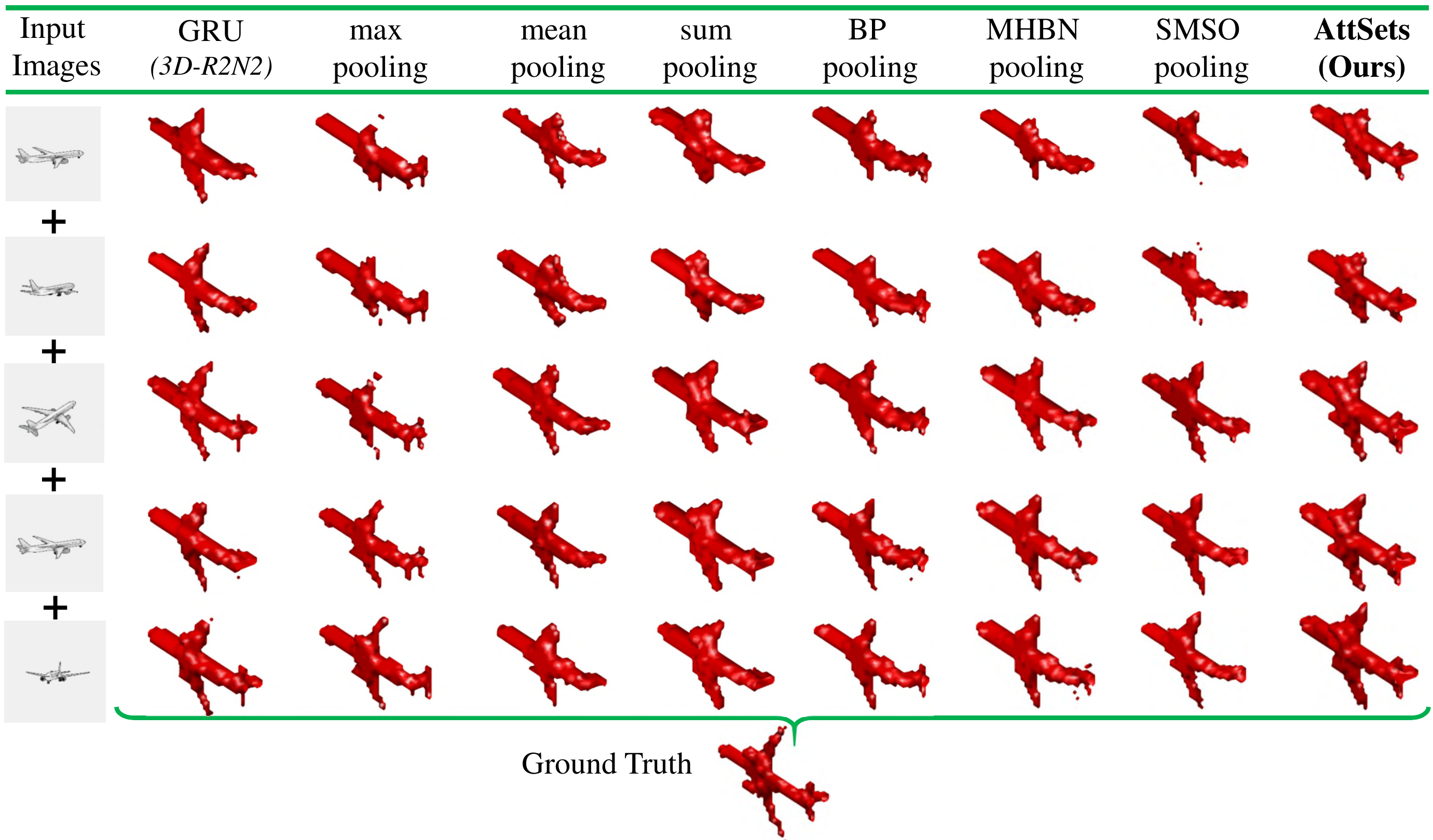}
\caption{Qualitative results of multi-view reconstruction from different approaches in ShapeNet$_{\textrm{lsm}}$ testing split.}
\label{fig:mv_demo_lsm}
\vspace{-0.35cm}
\end{figure*}

\begin{table*}[ht]
\caption{Mean IoU for multi-view reconstruction of all 13 categories from ShapeNet$_{\textrm{lsm}}$ dataset. All networks are well trained in previous experiment Group 5 of Section \ref{sec:eval_r2n2}.}
\centering
\label{tab:iou_lsm_allv}
\tabcolsep=0.2cm
\begin{tabular}{ l|ccccccccc}
\hline
&1 view&2 views&3 views& 4 views&5 views&8 views&12 views&16 views&20 views \\
\hline
Base$_{\textrm{r2n2}}$-GRU &0.390&0.428&0.444&0.454&0.459&0.467&0.470&0.471&0.472 \\
Base$_{\textrm{r2n2}}$-max pooling &0.276&0.388&0.433&0.459&0.473&0.497&0.510&0.515&0.518 \\
Base$_{\textrm{r2n2}}$-mean pooling &0.365&0.426&0.452&0.468&0.477&0.493&0.503&0.508&0.511 \\
Base$_{\textrm{r2n2}}$-sum pooling &0.363&0.421&0.445&0.459&0.466&0.481&0.492&0.499&0.503 \\
Base$_{\textrm{r2n2}}$-BP pooling &0.359&0.407&0.426&0.436&0.442&0.453&0.459&0.462&0.463 \\
Base$_{\textrm{r2n2}}$-MHBN pooling &0.365&0.403&0.418&0.427&0.431&0.441&0.446&0.449&0.450 \\
Base$_{\textrm{r2n2}}$-SMSO pooling &0.364&0.419&0.445&0.460&0.469&0.488&0.500&0.506&0.510 \\
\textbf{Base$_{\textrm{r2n2}}$-\nickname{}(Ours)} &\textbf{0.404}&\textbf{0.452}&\textbf{0.475}&\textbf{0.490}&\textbf{0.498}&\textbf{0.514}
&\textbf{0.522}&\textbf{0.528}&\textbf{0.531} \\
\hline
\end{tabular}
\vspace{-0.3 cm}
\end{table*}

\begin{table*}[t]
\caption{Group 1: mean IoU for multi-view reconstruction of all 40 categories in ModelNet40 testing split. All networks are firstly trained given only 1 image for each object in Stage 1. The \nickname{} module is further trained given \textbf{12 images} per object in Stage 2, while other competing approaches are fine-tuned given \textbf{12 images} per object in Stage 2.}
\centering
\label{tab:iou_modelnet_12v}
\tabcolsep=0.4cm
\begin{tabular}{ l|ccccccc}
\hline
&1 view&2 views&3 views& 4 views&5 views&8 views&12 views \\
\hline
Base$_{\textrm{r2n2}}$-GRU &0.344&0.390&0.414&0.430&0.440&0.454&0.464\\
Base$_{\textrm{r2n2}}$-max pooling &0.393&0.468&0.490&0.504&0.511&0.523&0.525\\
Base$_{\textrm{r2n2}}$-mean pooling &0.415&0.464&0.481&0.495&0.502&0.515&0.520\\
Base$_{\textrm{r2n2}}$-sum pooling &0.332&0.441&0.473&0.492&0.500&0.514&0.520\\
Base$_{\textrm{r2n2}}$-BP pooling &0.431&0.466&0.479&0.492&0.497&0.509&0.515\\
Base$_{\textrm{r2n2}}$-MHBN pooling &0.423&0.462&0.478&0.491&0.497&0.509&0.515\\
Base$_{\textrm{r2n2}}$-SMSO pooling &0.441&0.476&0.490&0.500&0.506&0.517&0.520\\
\textbf{Base$_{\textrm{r2n2}}$-\nickname{}(Ours)} &\textbf{0.487}&\textbf{0.505}&\textbf{0.511}&\textbf{0.517}&\textbf{0.521}&\textbf{0.527}
&\textbf{0.529} \\
\hline
\end{tabular}
\end{table*}

\begin{table*}[t]
\caption{Group 2: mean IoU for multi-view reconstruction of all 40 categories in ModelNet40 testing split. All networks are firstly trained given only 1 image for each object in Stage 1. The \nickname{} module is further trained given random number of images per object in Stage 2, \ie $N$ is uniformly sampled from \textbf{[1, 12]}, while other competing approaches are fine-tuned given random number of views per object in Stage 2.}
\centering
\label{tab:iou_modelnet_allv}
\tabcolsep=0.4cm
\begin{tabular}{ l|ccccccc}
\hline
&1 view&2 views&3 views& 4 views&5 views&8 views&12 views \\
\hline
Base$_{\textrm{r2n2}}$-GRU &0.388&0.421&0.434&0.440&0.444&0.449&0.452\\
Base$_{\textrm{r2n2}}$-max pooling &0.461&0.489&0.498&0.506&0.509&0.515&0.517\\
Base$_{\textrm{r2n2}}$-mean pooling &0.455&0.487&0.498&0.507&0.512&0.520&0.523\\
Base$_{\textrm{r2n2}}$-sum pooling &0.453&0.484&0.494&0.503&0.506&0.514&0.517\\
Base$_{\textrm{r2n2}}$-BP pooling &0.454&0.479&0.487&0.496&0.499&0.507&0.510\\
Base$_{\textrm{r2n2}}$-MHBN pooling &0.453&0.480&0.488&0.497&0.500&0.507&0.509\\
Base$_{\textrm{r2n2}}$-SMSO pooling &0.462&0.488&0.497&0.505&0.509&0.516&0.519\\
\textbf{Base$_{\textrm{r2n2}}$-\nickname{}(Ours)} &\textbf{0.487}&\textbf{0.505}&\textbf{0.511}&\textbf{0.518}&\textbf{0.520}&\textbf{0.525}
&\textbf{0.527} \\
\hline
\end{tabular}
\vspace{-0.1 cm}
\end{table*}
\begin{figure*}[t]
\centering
   \includegraphics[width=1\linewidth]{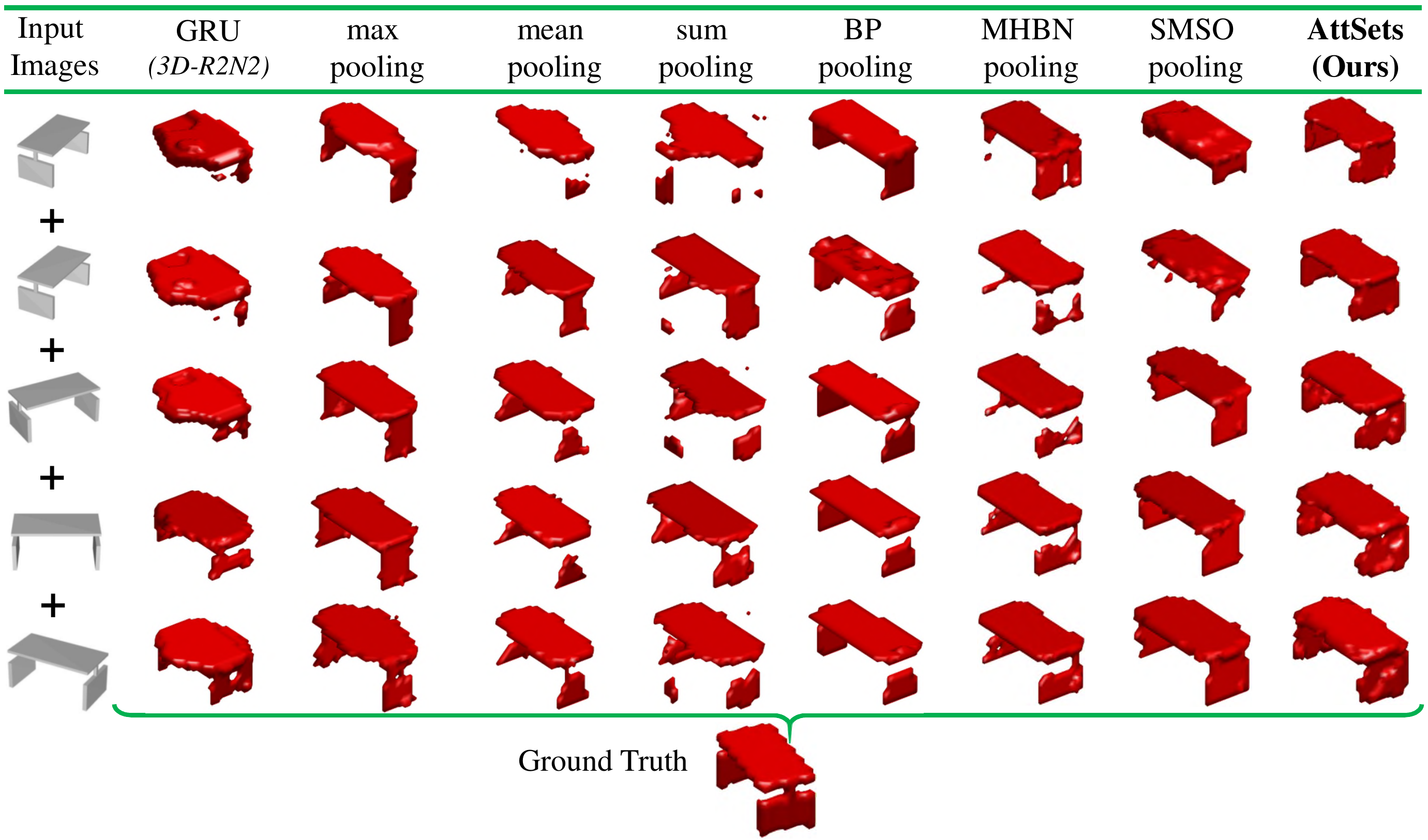}
\caption{Qualitative results of multi-view reconstruction from different approaches in ModelNet40 testing split.}
\label{fig:mv_demo_modelnet40}
\vspace{-0.4cm}
\end{figure*}

\textbf{Analysis.} We investigate the results as follows: 
\vspace{-0.2cm}
\begin{itemize}[leftmargin=0.3cm]
\item The GRU based approach can generate reasonable 3D shapes in all experiments Group 1 $\sim$ 5 given either few or multiple images during testing, but the performance saturates quickly after being given more images, \eg 8 views, because the recurrent unit is hardly able to capture features from longer image sequences, as illustrated in Figure \ref{fig:24v_mIoU} \circled{1}.

\item In Group 1 $\sim$ 4, all pooling based approaches are able to estimate satisfactory 3D shapes when given a similar number of images as given in training, but they are unlikely to predict reasonable shapes given an arbitrary number of images. For example, in experiment Group 4, all pooling based approaches have inferior IoU scores given only few images as shown in Table \ref{tab:iou_r2n2_24v} and Figure \ref{fig:24v_mIoU} \circled{2}, because the pooled features from fewer images during testing are unlikely to be as general and representative as pooled features from more images during training. Therefore, those models trained on $24$ images fail to generalize well to only one image during testing.

\item In Group 5, as shown in Table \ref{tab:iou_r2n2_allv} and Figure \ref{fig:allv_mIoU}, all pooling based approaches are much more robust compared with Group 1$\sim$4, because the networks are generally optimized according to an arbitrary number of images during training. However, these networks tend to have the performance \textit{in the middle}. Compared with Group 4, all approaches in Group 5 tend to have better performance when $N=1$, while being worse when $N=24$. Compared with Group 1, all approaches in Group 5 are likely to be better when $N=24$, while being worse when $N=1$. Basically, these networks tend to be optimized to learn the \textit{mean} features overall.

\item  In all experiments Group 1 $\sim$ 5, all approaches tend to have better performance when given enough input images, \ie $N=24$, because more images are able to provide enough information for reconstruction.

\item In all experiments Group 1 $\sim$ 5, our \nickname{} based approach clearly outperforms all others in either single or multiple view 3D reconstruction and it is more robust to a variable number of input images. Our \faset{} algorithm completely decouples the base network to learn visual features for accurate single view reconstruction as illustrated in Figure \ref{fig:24v_mIoU} \circled{3}, while the trainable parameters of \nickname{} module are separately responsible for learning attention scores for better multi-view reconstruction as shown in Figure \ref{fig:24v_mIoU} \circled{4}. Therefore, the whole network does not suffer from limitations of GRU or pooling approaches, and can achieve better performance for either fewer or more image reconstruction.
\end{itemize}

\vspace{-0.45cm}
\subsection{Evaluation on ShapeNet$_{\textrm{lsm}}$ Dataset}\label{sec:eval_lsm}
To further investigate how well the learnt visual features and attention scores generalize across different style of images, we use the well trained networks of previous Group 5 of Section \ref{sec:eval_r2n2} to test on the large ShapeNet$_{\textrm{lsm}}$ dataset. Note that, we only borrow the synthesized images from ShapeNet$_{\textrm{lsm}}$ dataset corresponding to the objects in ShapeNet$_{\textrm{r2n2}}$ testing split. This guarantees that all the trained models have never seen either the style of LSM rendered images or the 3D object labels before. The image viewing angles from the original ShapeNet$_{\textrm{lsm}}$ dataset are not used in our experiments, since the Base$_{\textrm{r2n2}}$ network does not require image viewing angles as input. Table \ref{tab:iou_lsm_allv} shows the mean IoU scores of all approaches, while Figure \ref{fig:mv_demo_lsm} shows the qualitative results. 

Our \nickname{} based approach outperforms all others given either few or multiple input images. This demonstrates that our Base$_{\textrm{r2n2}}$-AttSets approach does not overfit the training data, but has better generality and robustness over new styles of rendered color images compared with other approaches.

\vspace{-0.45cm}
\subsection{Evaluation on ModelNet40 Dataset}\label{sec:eval_modelnet40}
We train the Base$_{\textrm{r2n2}}$-AttSets and its competing approaches on ModelNet40 dataset from scratch. For fair comparison, all networks \rev{(the pooling/GRU/\nickname{} based approaches)} are trained according to the proposed \faset{} algorithm, which is similar to the two-stage training strategy of Section \ref{sec:eval_r2n2}. 

\textbf{Training Stage 1.} All networks are trained given only 1 image for each object, \ie $N=1$ in all training iterations, until convergence. This guarantees all networks are well optimized for single view 3D reconstruction.

\textbf{Training Stage 2.} We further conduct the following two parallel groups of training experiments to optimize the networks for multi-view reconstruction.
\begin{itemize}[leftmargin=0.3cm]
\item Group 1. All networks are further trained given all 12 images for each object, \ie $N=12$ in all iterations, until convergence. As to our Base$_{\textrm{r2n2}}$-AttSets, the well-trained encoder-decoder in previous Stage 1 is frozen, and only the \nickname{} module is trained. All other competing approaches are fine-tuned using smaller learning rate (1e-5) in this stage.

\item Group 2. All networks are further trained until convergence, but $N$ is uniformly and randomly sampled from $[1, 12]$ for each object during training. Only the \nickname{} module is trained, while all other competing approaches are fine-tuned in this Stage 2.
\end{itemize}

\begin{table}[ht]
\caption{Group 1: mean IoU for silhouettes prediction on the Blobby dataset. All networks are firstly trained given only 1 image for each object in Stage 1. The \nickname{} module is further trained given \textbf{2 images} per object, \ie $N$ =2, while other competing approaches are fine-tuned given 2 views per object in Stage 2.}
\centering
\label{tab:iou_blobby_02v}
\tabcolsep=0.05cm
\begin{tabular}{ l|cccc}
\hline
&1 view&2 views&3 views& 4 views \\
\hline
Base$_{\textrm{silnet}}$-GRU &0.857&0.860&0.860&0.860\\
Base$_{\textrm{silnet}}$-max pooling &0.922&0.923&0.924&0.924\\
Base$_{\textrm{silnet}}$-mean pooling &0.920&0.922&0.923&0.924\\
Base$_{\textrm{silnet}}$-sum pooling &0.913&0.918&0.917&0.916\\
Base$_{\textrm{silnet}}$-BP pooling &0.908&0.912&0.914&0.914\\
Base$_{\textrm{silnet}}$-MHBN pooling &0.901&0.904&0.906&0.906\\
Base$_{\textrm{silnet}}$-SMSO pooling &0.860&0.865&0.865&0.865\\
\textbf{Base$_{\textrm{silnet}}$-\nickname{}(Ours)} &\textbf{0.924}&\textbf{0.931}&\textbf{0.933}&\textbf{0.935} \\
\hline
\end{tabular}
\vspace{-0.35 cm}
\end{table}
\begin{table}[h]
\caption{Group 2: mean IoU for silhouettes prediction on the Blobby dataset. All networks are firstly trained given only 1 image for each object in Stage 1. The \nickname{} module is further trained given \textbf{4 images} per object, \ie $N$=4, while other competing approaches are fine-tuned given 4 views per object in Stage 2.}
\centering
\label{tab:iou_blobby_04v}
\tabcolsep=0.05cm
\begin{tabular}{ l|cccc}
\hline
&1 view&2 views&3 views& 4 views \\
\hline
Base$_{\textrm{silnet}}$-GRU &0.863&0.865&0.865&0.865\\
Base$_{\textrm{silnet}}$-max pooling &0.923&0.927&0.929&0.929\\
Base$_{\textrm{silnet}}$-mean pooling &0.923&0.925&0.927&0.927\\
Base$_{\textrm{silnet}}$-sum pooling &0.902&0.917&0.921&0.924\\
Base$_{\textrm{silnet}}$-BP pooling &0.911&0.916&0.919&0.920\\
Base$_{\textrm{silnet}}$-MHBN pooling &0.904&0.908&0.911&0.911\\
Base$_{\textrm{silnet}}$-SMSO pooling &0.863&0.865&0.865&0.865\\
\textbf{Base$_{\textrm{silnet}}$-\nickname{}(Ours)} &\textbf{0.924}&\textbf{0.932}&\textbf{0.936}&\textbf{0.937} \\
\hline
\end{tabular}
\vspace{-0.35 cm}
\end{table}

\textbf{Testing Stage.} All networks trained in above two groups are separately tested given $N=[1,2,3,4,5,8,12]$. The permutations of input images are the same for all different approaches for fair comparison.

\textbf{Results.} Tables \ref{tab:iou_modelnet_12v} and \ref{tab:iou_modelnet_allv} show the mean IoU scores of Groups 1 and 2 respectively, and Figure \ref{fig:mv_demo_modelnet40} shows qualitative results of Group 2. The Base$_{\textrm{r2n2}}$-AttSets surpasses all competing approaches by a large margin for both single and multiple view 3D reconstruction, and all the results are consistent with previous experimental results on both ShapeNet$_{\textrm{r2n2}}$ and ShapeNet$_{\textrm{lsm}}$ datasets. 

\vspace{-0.45cm}
\subsection{Evaluation on Blobby Dataset}\label{sec:eval_blobby}
\begin{figure*}[t]
\centering
   \includegraphics[width=1\linewidth]{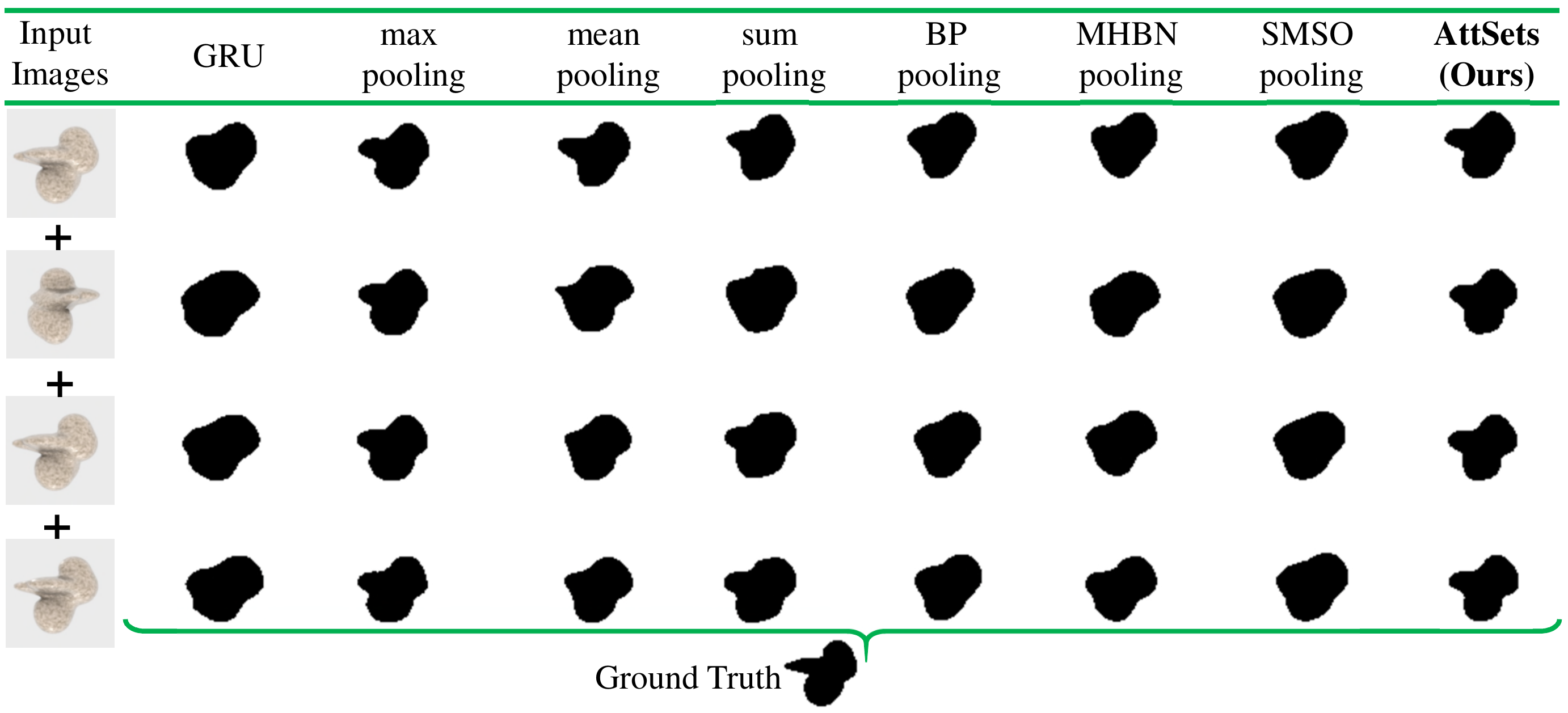}
\caption{Qualitative results of silhouettes prediction from different approaches on the Blobby dataset.}
\label{fig:mv_demo_blobby}
\vspace{-0.1cm}
\end{figure*}
In this section, we evaluate the Base$_{\textrm{silnet}}$-AttSets and the competing approaches on the Blobby dataset. For fair comparison, the GRU module is implemented with a single fully connected layer of 160 hidden units, which has similar network capacity with our \nickname{} based network. All networks \rev{(the pooling/GRU/\nickname{} based approaches)} are trained with the proposed two-stage \faset{} algorithm as follows:

\textbf{Training Stage 1.} All networks are trained given only 1 image together with the viewing angle for each object, \ie $N$=1 in all training iterations, until convergence. This guarantees the performance of single view shape learning. 

\textbf{Training Stage 2.} Another two parallel groups of training experiments are conducted to further optimize the networks for multi-view shape learning.
\vspace{-0.2cm}
\begin{itemize}[leftmargin=0.2cm]
\item Group 1. All networks are further trained given only 2 images for each object, \ie $N$=2 in all iterations. As to Base$_{\textrm{silnet}}$-AttSets, only the \nickname{} module is optimized with the well-trained base encoder-decoder being frozen. For fair comparison, all competing approaches are fine-tuned given 2 images per object for better performance where $N$ =2 until convergence.
\item Group 2. Similar to the above Group 1, all networks are further trained given all 4 images for each object, \ie $N$=4, until convergence.
\end{itemize}

\textbf{Testing Stage.} All networks trained in above two groups are separately tested given $N$ = [1,2,3,4]. The permutations of input images are the same for all different networks for fair comparison.

\textbf{Results.} Table \ref{tab:iou_blobby_02v} and \ref{tab:iou_blobby_04v} show the mean IoUs of above two groups of experiments and Figure \ref{fig:mv_demo_blobby} shows the qualitative results of Group 2. Note that, the IoUs are calculated on predicted 2D silhouettes instead of 3D voxels, so they are not numerically comparable with previous experiments on ShapeNet$_{\textrm{r2n2}}$, ShapeNet$_{\textrm{lsm}}$, and ModelNet40 datasets. We do not include the IoU scores of the original SilNet \citep{Wiles2017}, because the original IoU scores are obtained from an end-to-end training strategy. In this paper, we uniformly apply the proposed two-stage \faset{} training paradigm on all approaches for fair comparison. Our Base$_{\textrm{silnet}}$-AttSets consistently outperforms all competing approaches for shape learning from either single or multiple views.

\begin{figure*}[t]
\centering
   \includegraphics[width=1\linewidth]{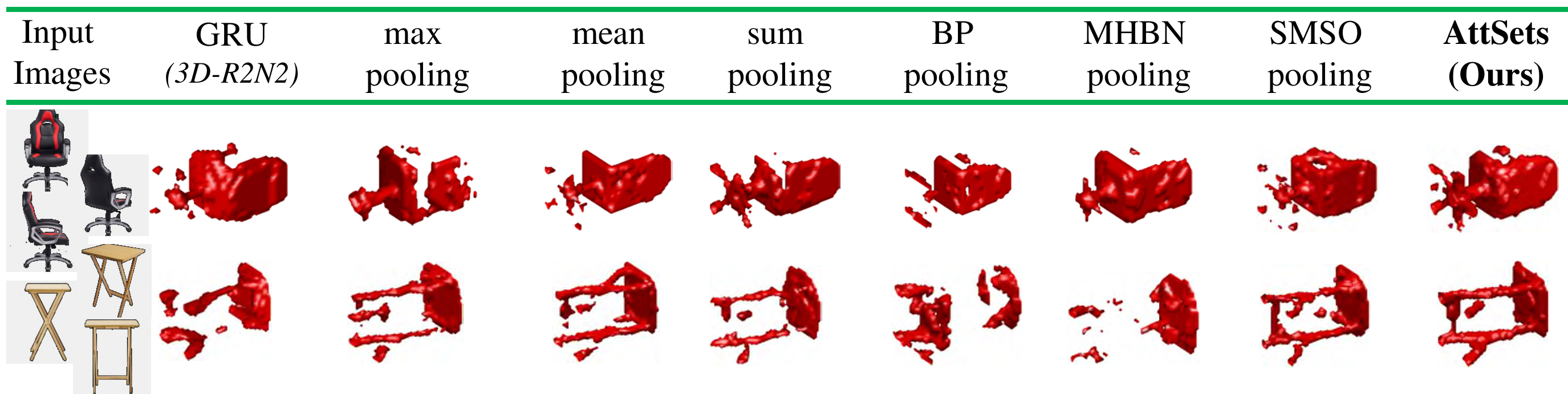}
\caption{Qualitative results of multi-view 3D reconstruction from real-world images.}
\label{fig:mv_demo_real}
\vspace{-0.3cm}
\end{figure*}
\begin{figure}[t]
\centering
   \includegraphics[width=1\linewidth]{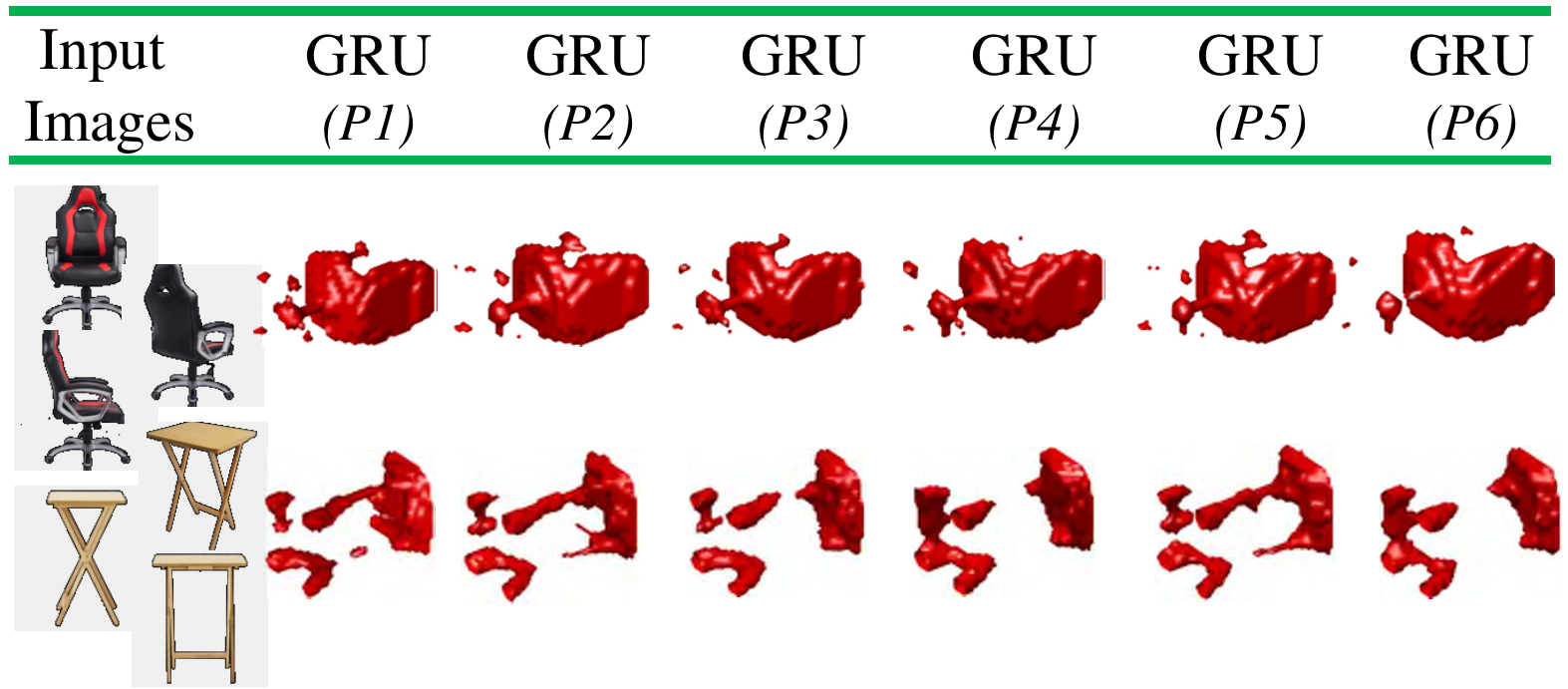}
\caption{Qualitative results of inconsistent 3D reconstruction from the GRU based approach.}
\label{fig:mv_demo_real_permu}
\vspace{-0.3cm}
\end{figure}

\vspace{-0.6cm}
\subsection{Qualitative Results on Real-world Images}
To the best of our knowledge, there is no public real-world dataset for multi-view 3D object reconstruction. Therefore, we manually collect real world images from Amazon online shops to qualitatively demonstrate the generality of all networks which are trained on the synthetic ShapeNet$_{\textrm{r2n2}}$ dataset in experiment Group 4 of Section \ref{sec:eval_r2n2}, as shown in Figure \ref{fig:mv_demo_real}.

In the meantime, we use these real-world images to qualitatively show the permutation invariance of different approaches. In particular, for each object, we use 6 different permutations in total for testing. As shown in Figure \ref{fig:mv_demo_real_permu}, the GRU based approach generates inconsistent 3D shapes given different image permutations. For example, the arm of a chair and the leg of a table can be reconstructed in permutation 1, but fail to be recovered in another permutation. By comparison, all other approaches are permutation invariant, as the results shown in Figure \ref{fig:mv_demo_real}.

\vspace{-0.75cm}
\subsection{Computational Efficiency}
To evaluate the computation and memory cost of \nickname{}, we implement Base$_{\textrm{r2n2}}$-AttSets and the competing approaches in Python 2.7 and Tensorflow 1.2 with CUDA 9.0 and cuDNN 7.1 as the back-end driver and library. All approaches share the same Base$_{\textrm{r2n2}}$ network and run in the same Titan X and software environments. Table \ref{tab:time_con} shows the average time consumption to reconstruct a single 3D object given different number of images. Our \nickname{} based approach is as efficient as the pooling methods, while Base$_{\textrm{r2n2}}$-GRU (\ie 3D-R2N2) takes more time when processing an increasing number of images due to the sequential computation mechanism of its GRU module. In terms of the total trainable weights, the max/mean/sum pooling based approaches have $16.66$ million, while \nickname{} based net has $17.71$ million. By contrast, the original 3D-R2N2 has $34.78$ million, the BP/MHBN/SMSO have $141.57, 60.78$ and $17.71$ million respectively. Overall, our \nickname{} outperforms the recurrent unit and pooling operations without incurring notable computation and memory cost.

\begin{table}[t]
\caption{Mean time consumption for a single object ($32^3$ voxel grid) estimation from different number of images (milliseconds).}
\centering
\label{tab:time_con}
\tabcolsep=0.02cm
\begin{tabular}{ l|ccccccc}
\hline
number of input images&1 &4 &8 &12 &16 &20 &24  \\
\hline
Base$_{\textrm{r2n2}}$-GRU &6.9&11.2&17.0&22.8&28.8&34.7&40.7\\
Base$_{\textrm{r2n2}}$-max pooling &6.4&\textbf{10.0}&\textbf{15.1}&20.2&\textbf{25.3}&\textbf{30.2}&\textbf{35.4}\\
Base$_{\textrm{r2n2}}$-mean pooling &\textbf{6.3}&10.1&\textbf{15.1}&\textbf{20.1}&\textbf{25.3}&30.3&35.5 \\
Base$_{\textrm{r2n2}}$-sum pooling &6.4&10.1&\textbf{15.1}&\textbf{20.1}&\textbf{25.3}&30.3&35.5 \\
Base$_{\textrm{r2n2}}$-BP pooling &6.5 &10.5 & 15.6 & 20.5&25.7& 30.6&35.8 \\
Base$_{\textrm{r2n2}}$-MHBN pooling &6.5 &10.3 & 15.3 & 20.3&25.5& 30.7&35.7 \\
Base$_{\textrm{r2n2}}$-SMSO pooling &6.5 &10.2 & 15.3 & 20.3&25.4&30.5&35.6 \\
\textbf{Base$_{\textrm{r2n2}}$-\nickname{}(Ours)} &7.7&11.0&16.3&21.2&26.3&31.4&36.4\\
\hline
\end{tabular}
\vspace{-0.2cm}
\end{table}

\begin{table*}[t]
\caption{ \small{Mean IoU of \nickname{} variants on all 13 categories in ShapeNet$_{\textrm{r2n2}}$ testing split.}}
\centering
\label{tab:iou_variants}
\tabcolsep=0.112cm
\begin{tabular}{ l|cccccccccc}
\hline
&1 view&2 views&3 views& 4 views&5 views&8 views&12 views&16 views&20 views&24 views \\
\hline
\textbf{Base$_{\textrm{r2n2}}$-\nickname{} ($conv2d$)}&\textbf{0.642}&0.648&0.651&0.655&0.657&0.664&0.668&0.674&0.675&0.676\\
\textbf{Base$_{\textrm{r2n2}}$-\nickname{} ($conv3d$)}&\textbf{0.642}&\textbf{0.663}&\textbf{0.671}&\textbf{0.676}&\textbf{0.677}&0.683&0.685&0.689&0.690&0.690 \\
\textbf{Base$_{\textrm{r2n2}}$-\nickname{} ($fc$)} &\textbf{0.642}&0.660&0.668&0.674&0.676&\textbf{0.684}
&\textbf{0.688}&\textbf{0.693}&\textbf{0.694}&\textbf{0.695} \\
\hline
\end{tabular}
\vspace{-0.1cm}
\end{table*}

\begin{table*}[t]
\caption{ \small{Mean IoU of all 13 categories in ShapeNet$_{\textrm{r2n2}}$ testing split for feature-wise and element-wise attentional aggregation.}}
\centering
\label{tab:iou_fw_ew}
\tabcolsep=0.078cm
\begin{tabular}{ l|cccccccccc}
\hline
&1 view&2 views&3 views& 4 views&5 views&8 views&12 views&16 views&20 views&24 views \\
\hline
\textbf{Base$_{\textrm{r2n2}}$-\nickname{}} \scriptsize{(element-wise)} &\textbf{0.642}&0.653&0.657&0.660&0.661&0.665&0.667&0.670&0.671&0.672\\
\textbf{Base$_{\textrm{r2n2}}$-\nickname{} \scriptsize{(feature-wise)}} &\textbf{0.642}&\textbf{0.660}&\textbf{0.668}&\textbf{0.674}&\textbf{0.676}&\textbf{0.684}
&\textbf{0.688}&\textbf{0.693}&\textbf{0.694}&\textbf{0.695} \\
\hline
\end{tabular}
\vspace{-0.1cm}
\end{table*}

\begin{table*}[t]
\caption{ \small{Mean IoU of different training algorithms on all 13 categories in ShapeNet$_{\textrm{r2n2}}$ testing split.}}
\centering
\label{tab:iou_train_alg}
\tabcolsep=0.11cm
\begin{tabular}{ l|cccccccccc}
\hline
&1 view&2 views&3 views& 4 views&5 views&8 views&12 views&16 views&20 views&24 views \\
\hline
\textbf{Base$_{\textrm{r2n2}}$-\nickname{}} (JoinT)&0.307&0.437&0.516&0.563&0.595&0.639&0.659&0.673&0.677&0.680\\
\textbf{Base$_{\textrm{r2n2}}$-\nickname{} (\faset{})} &\textbf{0.642}&\textbf{0.660}&\textbf{0.668}&\textbf{0.674}&\textbf{0.676}&\textbf{0.684}
&\textbf{0.688}&\textbf{0.693}&\textbf{0.694}&\textbf{0.695} \\
\hline
\end{tabular}
\vspace{-0.3cm}
\end{table*}

\vspace{-0.45cm}
\subsection{Comparison between Variants of \nickname{}}\label{sec:variants}
\begin{figure}
\vspace{-0.15cm}
\centering
   \includegraphics[width=1\linewidth]{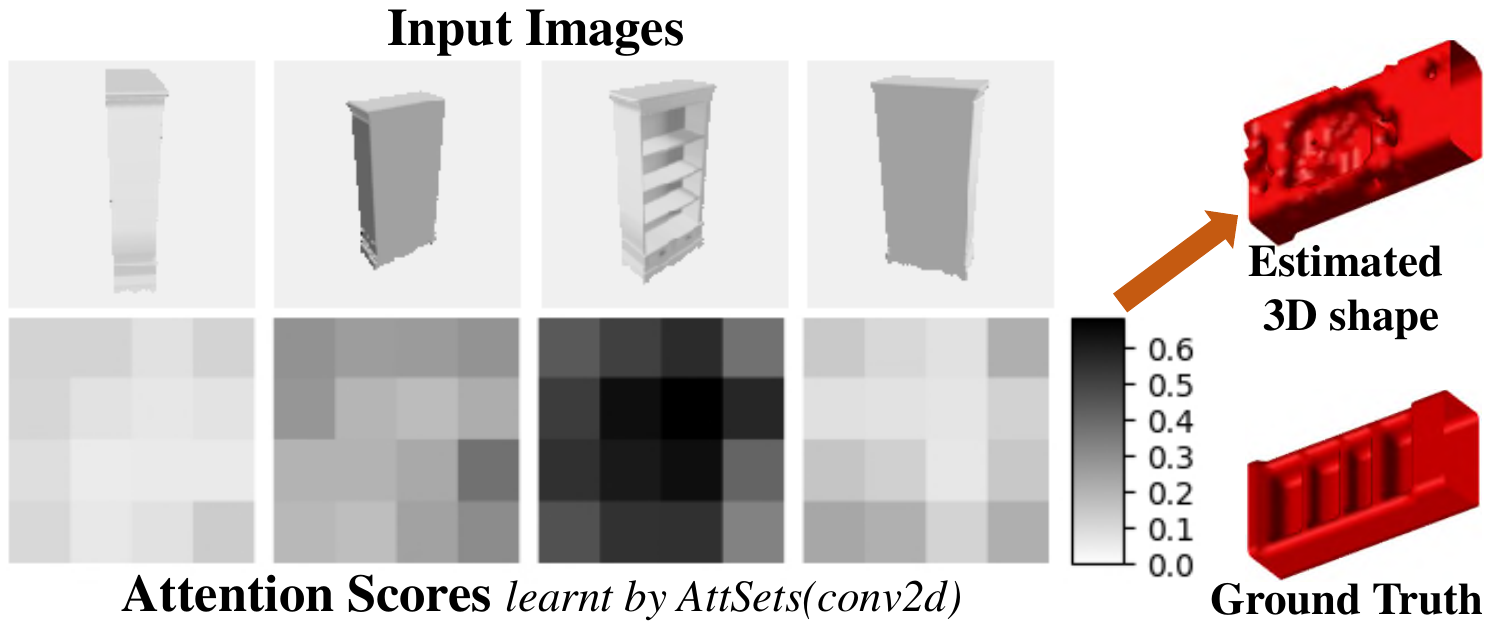}
\caption{Learnt attention scores for deep feature sets via $conv2d$ based \nickname{}.}
\label{fig:atts_w}
\vspace{-0.35cm}
\end{figure}

We further compare the aggregation performance of $fc$, $conv2d$ and $conv3d$ based \nickname{} variants which are shown in Figure \ref{fig:attsets_imp} in Section \ref{sec:impl}. The $fc$ based \nickname{} net is the same as in Section \ref{sec:eval_r2n2}. The $conv2d$ based \nickname{} is plugged into the middle of the 2D encoder, fusing a $(N, 4, 4, 256)$ tensor into $(1, 4, 4, 256)$, where $N$ is an arbitrary image number. The $conv3d$ based \nickname{} is plugged into the middle of the 3D decoder, integrating a $(N, 8, 8, 8, 128)$ tensor into $(1, 8, 8, 8, 128)$. All other layers of these variants are the same. Both the $conv2d$ and $conv3d$ based \nickname{} networks are trained using the paradigm of experiment Group 4 in Section \ref{sec:eval_r2n2}.
Table \ref{tab:iou_variants} shows the mean IoU scores of three variants on ShapeNet$_{\textrm{r2n2}}$ testing split. $fc$ and $conv3d$ based variants achieve similar IoU scores for either single or multi view 3D reconstruction, demonstrating the superior aggregation capability of \nickname{}. In the meantime, we observe that the overall performance of $conv2d$ based \nickname{} net is slightly decreased compared with the other two. One possible reason is that the 2D feature set has been aggregated at the early layer of the network, resulting in features being lost early. Figure \ref{fig:atts_w} visualizes the learnt attention scores for a 2D feature set, \ie $(N,4,4,256)$ features, via the $conv2d$ based \nickname{} net. To visualize 2D feature scores, we average the scores along the channel axis and then roughly trace back the spatial locations of those scores corresponding to the original input. The more visual information the input image has, the higher attention scores are learnt by \nickname{} for the corresponding latent features. For example, the third image has richer visual information than the first image, so its attention scores are higher. Note that, for a specific base network, there are many potential locations to plug in \nickname{} and it is also possible to include multiple \nickname{} modules into the same net. To fully evaluate these factors is out of the scope of this paper.

\vspace{-0.45cm}
\subsection{Feature-wise Attention $vs.$ Element-wise Attention}
\begin{figure}
\vspace{-0.15cm}
\centering
   \includegraphics[width=1\linewidth]{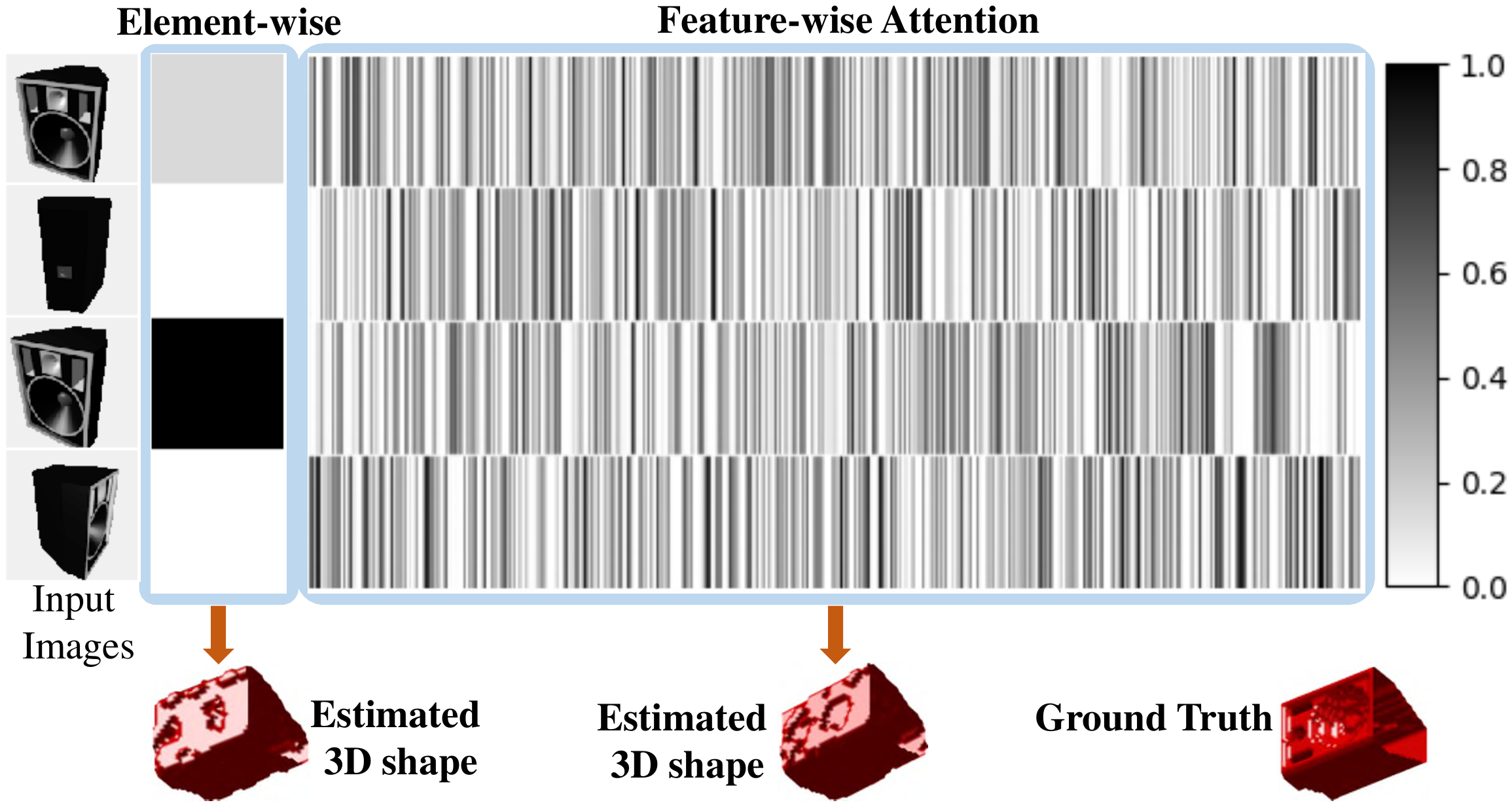}
\caption{Learnt attention scores for deep feature sets via element-wise attention and feature-wise attention \nickname{}.}
\label{fig:atts_ele}
\vspace{-0.5cm}
\end{figure}

Our \nickname{} module is initially designed to learn unique feature-wise attention scores for the whole input deep feature set, and we demonstrate that it significantly improves the aggregation performance over dynamic feature sets in previous Section \ref{sec:eval_r2n2}, \ref{sec:eval_lsm}, \ref{sec:eval_modelnet40} and \ref{sec:eval_blobby}. In this section, we further investigate the advantage of this feature-wise attentive pooling over element-wise attentional aggregation.

For element-wise attentional aggregation, the \nickname{} module turns to learn a single attention score for each element of the feature set $\mathcal{A} = \{\boldsymbol{x}_1, \boldsymbol{x}_2, \cdots, \boldsymbol{x}_N\}$, followed by the $softmax$ normalization and weighted summation pooling. In particular, as shown in previous Figure \ref{fig:attsets_f}, the shared function $g(\boldsymbol{x}_n, \boldsymbol{W})$ now learns a scalar, instead of a vector, as the attention activation for each input element. Eventually, all features within the same element are weighted by a learnt common attention score. Intuitively, the original feature-wise \nickname{} tends to be fine-grained aggregation, while the element-wise \nickname{} learns to coarsely aggregate features.

Following the same training settings of experiment Group 4 in Section \ref{sec:eval_r2n2}, we conduct another group of experiment on ShapeNet$_{\textrm{r2n2}}$ dataset for element-wise attentional aggregation. Table \ref{tab:iou_fw_ew} compares the mean IoU for 3D object reconstruction through feature-wise and element-wise attentional aggregation. Figure \ref{fig:atts_ele} shows an example of the learnt attention scores and the predicted 3D shapes. As expected, the feature-wise attention mechanism clearly achieves better aggregation performance compared with the coarsely element-wise approach. As shown in Figure \ref{fig:atts_ele}, the element-wise attention mechanism tends to focus on few images, while completely ignoring others. By comparison, the feature-wise \nickname{} learns to fuse information from all images, thus achieving better aggregation performance.

\vspace{-0.3cm}
\subsection{Significance of \faset{} Algorithm}\label{sec:sig_faset}
In this section, we investigate the impact of \faset{} algorithm by comparing it with the standard end-to-end joint training (JoinT). Particularly, in JoinT, all parameters $\Theta_{base}$ and $\Theta_{att}$ are jointly optimized with a single loss. Following the same training settings of experiment Group 4 in Section \ref{sec:eval_r2n2}, we conduct another group of experiment on ShapeNet$_{\textrm{r2n2}}$ dataset under the JoinT training strategy. As its IoU scores shown in Table \ref{tab:iou_train_alg}, the JoinT training approach tends to optimize the whole net regarding the training multi-view batches, thus being unable to generalize well for fewer images during testing. Basically, the network itself is unable to dedicate the base layers to learning visual features, while the \nickname{} module to learning attention scores, if it is not trained with the proposed \faset{} algorithm. The theoretical reason is discussed previously in Section \ref{sec:optim_motiv}. \rev{The \faset{} algorithm may also be applicable to other learning based aggregation approaches, as long as the aggregation module can be decoupled from the base encoder/decoder.}

\vspace{-0.4cm}
\section{Conclusion}
In this paper, we present \nickname{} module and \faset{} training algorithm to aggregate elements of deep feature sets. \nickname{} together with \faset{} has powerful permutation invariance, computation efficiency, robustness and flexible implementation properties, along with the theory and extensive experiments to support its performance for multi-view 3D reconstruction. Both quantitative and qualitative results explicitly show that \nickname{} significantly outperforms other widely used aggregation approaches. Nevertheless, all of our experiments are dedicated to multi-view 3D reconstruction. It would be interesting to explore the generality of \nickname{} and \faset{} over other set-based tasks, especially the tasks which constantly take multiple elements as input.

\clearpage
\bibliographystyle{spbasic}      
\bibliography{Mendeley}   

\end{document}